\documentclass{article}

\PassOptionsToPackage{numbers}{natbib}
\usepackage{amsmath,amsfonts,bm}

\def\eqref#1{equation~\ref{#1}}
\def\1{\bm{1}}

\DeclareMathAlphabet{\mathsfit}{\encodingdefault}{\sfdefault}{m}{sl}
\SetMathAlphabet{\mathsfit}{bold}{\encodingdefault}{\sfdefault}{bx}{n}

\usepackage[final]{neurips_2021}

\usepackage[utf8]{inputenc} % allow utf-8 input
\usepackage[T1]{fontenc}    % use 8-bit T1 fonts
\usepackage{hyperref}       % hyperlinks
\usepackage{url}            % simple URL typesetting
\usepackage{booktabs}       % professional-quality tables
\usepackage{amsfonts}       % blackboard math symbols
\usepackage{amsmath}
\usepackage{caption}
\usepackage{nicefrac}       % compact symbols for 1/2, etc.
\usepackage{microtype}      % microtypography
\usepackage{xcolor}         % colors
\usepackage{graphicx}
\usepackage{bbm}
\usepackage{booktabs}
\usepackage{mathtools}
\usepackage[inline]{enumitem}
\usepackage[ruled,vlined,nofillcomment]{algorithm2e}
\usepackage{multirow}
\usepackage{wrapfig}
\usepackage{booktabs}
\usepackage{listings}
\usepackage{color}
\usepackage{comment}
\newcommand{\std}[1]{$^{\pm #1}$}

\newcommand{\Li}{L^{\mathrm{in}}}
\newcommand{\Lo}{L^{\mathrm{out}}}

\SetKwInput{KwRequire}{Require}

\title{Learning where to learn:\\Gradient sparsity in meta and continual learning}

\author{%
\textbf{Johannes von Oswald*\textsuperscript{,1}, Dominic Zhao*\textsuperscript{,1},}\\
\textbf{Seijin Kobayashi\textsuperscript{1}, Simon Schug\textsuperscript{1}, Massimo Caccia\textsuperscript{2}, }\\
  \textbf{Nicolas Zucchet\textsuperscript{1}, João Sacramento\textsuperscript{1}}\\
  \\
  *Equal contribution\\
  \\
  \textsuperscript{1}Institute of Neuroinformatics, University of Zürich and ETH Zürich\\
  \textsuperscript{2}Mila, University of Montreal \& ServiceNow\\
  \texttt{\{voswaldj,dozhao\}@ethz.ch}  
}

\begin{document}

\maketitle

\begin{abstract}
Finding neural network weights that generalize well from small datasets is difficult. A promising approach is to learn a weight initialization such that a small number of weight changes results in low generalization error. We show that this form of meta-learning can be improved by letting the learning algorithm decide which weights to change, i.e., by learning where to learn. We find that patterned sparsity emerges from this process, with the pattern of sparsity varying on a problem-by-problem basis. This selective sparsity results in better generalization and less interference in a range of few-shot and continual learning problems. Moreover, we find that sparse learning also emerges in a more expressive model where learning rates are meta-learned. Our results shed light on an ongoing debate on whether meta-learning can discover adaptable features and suggest that learning by sparse gradient descent is a powerful inductive bias for meta-learning systems.
\end{abstract}

\section{Introduction}
Meta-learning holds the promise of discovering inductive biases that improve the performance of a primary learning process.
Such a set of assumptions can materialize in various elements of the learner.
The well-known model-agnostic meta-learning \citep[MAML;][]{finn_model-agnostic_2017} algorithm aims to learn a neural network initialization that generalizes well to new learning tasks.
More sophisticated meta-learners augment this procedure by additionally modulating the inner-loop learning dynamics \citep{DBLP:journals/corr/LiZCL17,lee2018gradient, pmlr-v97-zintgraf19a, flennerhag2020metalearning, zhao_meta_learning_hypernetworks, DBLP:journals/corr/abs-1909-05557}.

It has been recently shown that applying MAML while adapting only last-layer weights leads to almost no decrease in performance in standard few-shot learning benchmarks. Our study builds upon the surprising effectiveness of this form of meta-learning, known as almost no inner-loop training \citep[ANIL;][]{raghu_rapid_2020}. Here, instead of deciding which weights to freeze a priori, we endow the meta-learner with the possibility to explicitly stop changing certain weights in the inner-loop learning process. We do this by introducing an adjustable binary mask which is elementwise multiplied with gradient updates. This can be understood as a simple form of learned gradient modulation that induces sparsity. Overfitting can thus be prevented and learning sped up by focusing adaptation to a sparse parameter subset, discovered by meta-learning.

We find that our sparse-MAML algorithm recovers a behavior that is reminiscent of ANIL. It induces high gradient sparsity in earlier layers of the network while allowing for adaptation in deeper layers including the network's output. Despite this reduction in the number of adaptable parameters, the sparse learning patterns formed by sparse-MAML do not overly specialize to the family of tasks observed during meta-learning; both ANIL and MAML are outperformed by the resulting sparse learners in cross-adaptation problems involving a shift in task distribution \cite[][]{chen2019closer,oh2021boil}. Furthermore, sparsity adapts intuitively to the number of inner-loop gradient steps as well as its learning rate, few-shot dataset size, and network specifications. This leads to a robust and interpretable variant of MAML that improves generalization by self-regularizing the parameters that the model should learn.

An exciting avenue of meta-learning research concerns continual learning. Learning tasks sequentially by gradient descent generally leads to poor results, as past tasks tend to be rapidly forgotten due to interfering weight updates. Such interference can be reduced with online meta-learning methods which optimize the base learning algorithm using both present and past data, kept in a replay buffer \citep{riemer2019learning,gupta_-maml_2020}. Our findings translate to this setting. We analyze the state-of-the-art look-ahead MAML algorithm \citep[La-MAML;][]{gupta_-maml_2020} which introduces per-parameter meta-learned learning rates and find that sparse learning emerges, as a large fraction of learning rates drops to zero. Notably, similarly high performance can be reached when meta-learning binary gradient masks only. Moreover, performance improves after endowing a version of MAML adapted for online learning \citep[][]{caccia2020online} with binary gradient masks. Thus, sparse learning can improve generalization, accelerate future learning, and reduce forgetting, and these benefits can be realized within online meta-learning.

\section{From MAML to sparse-MAML}
\label{sec:MAML-to-sp-MAML}
\paragraph{MAML.} The MAML algorithm seeks neural network weights $\theta$ from which only a few gradient descent steps suffice to reach high performance on a given task $\tau$, that is assumed to be drawn from a certain distribution $p(\tau)$. Formally, a task is defined by an outer loss function $\Lo_\tau$ and an inner loss function $\Li_\tau$. We will later make explicit the form the two loss functions can take depending on the problem being solved. The result of the inner loss  minimization is evaluated by the outer loss leading to the following optimization problem:
\begin{equation}
    \label{eq:MAML-problem}
    \min_\theta \mathbb{E}_{\tau\sim p(\tau)}\!\left[\Lo_\tau\!\left(\phi_{\tau,K}(\theta)\right)\right]
    ~\text{s.t.}~ \phi_{\tau,k+1} = \phi_{\tau,k} - \alpha \, \nabla_\phi \Li_\tau\!\left( \phi_{\tau,k} \right) ~ \text{and} ~ \phi_{\tau,0}=\theta,
\end{equation}
with $\alpha$ the inner-loop learning rate and $K$ the number of gradient descent steps. The initialization $\theta$ is then obtained by iterative updating, using
\begin{equation}
    \label{eqn:maml-theta-update}
    \theta \gets \theta - \gamma_\theta \, \mathbb{E}_{\tau\sim p(\tau)}\!\left[\text{d}_\theta \, \Lo_\tau\!\left(\phi_{\tau,K}(\theta)\right)\right]\!,
\end{equation}
with $\gamma_\theta$ the outer-loop learning rate. Note that we need the total derivative $\mathrm{d}_\theta$ in Eq.~\ref{eqn:maml-theta-update} and not the partial derivative $\nabla_\theta$ due to the complex relationship between $\phi_{\tau, k}$ and $\theta$. In practice, the expectations over the task distribution that appear above are estimated by Monte Carlo integration. The updates in $\theta$ therefore correspond to stochastic gradient descent on the expected outer loss.

In MAML, the total derivative w.r.t.~to $\theta$ is obtained by backpropagating through the inner optimization, a resource-intensive procedure. First-order MAML (FOMAML) drastically reduces the computational cost by setting to zero the second-order derivatives that appear when differentiating the inner-loop update.

\paragraph{Learning the learning rates.} Some variants of MAML focus on learning the learning rate and consider inner-loop updates of the following form:
    \begin{equation}
    \label{eq:phi_updates}
    \phi_{\tau,k+1} = \phi_{\tau,k} - \alpha \, M \, \nabla_\phi \Li_\tau\!\left(\phi_{\tau,k}\right)\!,
\end{equation}
for some learnable preconditioning matrix $M$, that is optimized similarly to the initialization $\theta$. Through $M$, these algorithms learn some information on the geometry of the loss with the hope of faster inner-loop optimization. Meta-SGD \citep{DBLP:journals/corr/LiZCL17} considers a diagonal $M$, i.e., learnable learning rates, meta-curvature \citep{DBLP:conf/nips/ParkO19} considers a block matrix, while vanilla MAML corresponds to the $M=\text{Id}$ case.

\paragraph{Sparse-MAML.} In line with these approaches, we introduce sparse-MAML. Together with an initial set of weights $\theta$, our algorithm dynamically learns the parameters which will be updated and the ones that will not. Hence, sparse-MAML learns where to learn. To do so, we use a vector $m$ (instead of a matrix $M$) that modulates the gradient in the inner-loop update in the following way:
\begin{equation}
\label{eq:sp-MAML-inner-loop}
    \phi_{\tau,k+1} = \phi_{\tau,k} - \alpha \! \left ( \mathbbm{1}_{m\geq 0} \circ \nabla_\phi \Li_\tau\!\left(\phi_{\tau,k}\right) \right )\!,
\end{equation}
with $\mathbbm{1}_{\cdot\geq 0}: \mathbb{R}^n \to \{0,1\}^n$ the step function that is applied elementwise to the underlying parameter vector $m \in \mathbb{R}^n$ and $\circ$ the pointwise multiplication. We differentiate the step function by considering it linear: this method is called the straight-through estimator \citep{bengio_estimating_2013} and it was recently used for similar large-scale masking \citep{ramanujan_whats_2020}. Following FOMAML, we ignore second-order derivatives. This leads to the update
\begin{equation}
    \label{eq:sp-MAML-mask-update}
    m \gets m + \alpha \, \gamma_m \, \mathbb{E}_{\tau \sim p(\tau)} \!\left [ \nabla_\phi \Lo_\tau\!\left(\phi_{\tau,K}\right) \circ \sum_{k=0}^{K-1} \nabla_\phi\Li_\tau\!\left( \phi_{\tau,k}\right) \right ]\!.
\end{equation}
A detailed derivation of the mask update, alongside the presentation of the initialization update, can be found in the supplementary material (SM).

Our mask update depends on the alignment between the outer-loss gradient $g_\tau^\text{out} \coloneqq \nabla_\phi \Lo_\tau\!\left(\phi_{\tau,K}\right)$ and the inner loss gradient $\overline{g}_\tau^\text{\,in} \coloneqq \sum_{k=0}^{K-1} \nabla_\phi\Li_\tau\!\left( \phi_{k}\right)$ accumulated over the inner loop trajectory. Learning tends to be shut off on coordinates $i$ for which these two quantities are of opposing sign, $\mathbb{E}_\tau \!\left [ g_{\tau,i}^\text{out} \, \overline{g}_{\tau,i}^\text{\,in} \right] < 0$. Such freezing of learning when parameter updates are conflicting on the training and validation sets can decrease negative interference across tasks, which can in turn improve generalization performance  \citep{nichol_first-order_2018,riemer2019learning}.

\section{Few-shot learning}

Finding a network that performs well when trained on few samples of unseen data can be formulated as a meta-learning problem. We study here the supervised few-shot learning setting where tasks comprise small labelled datasets. A loss function $\mathcal{L}(\phi, \mathcal{D})$ measures how much the predictions of a network parameterized by $\phi$ deviate from the ground truth labels on dataset $\mathcal{D}$. During meta-learning, the data of a given task $\tau$ is split into training and validation datasets, $\mathcal{D}_\tau^\text{t}$ and $\mathcal{D}_\tau^\text{v}$, respectively. The sparse-MAML formulation of few-shot learning then consists in optimizing the meta-parameters $\theta$ and $m$ that, given the training set, in turn yield parameters $\phi$ that improve validation set performance:
\begin{equation}
    \label{eq:supervised-few-shot}
    \begin{split}
        & \min_\theta \; \mathbb{E}_{\tau\sim p(\tau)}\!\left[\mathcal{L}\!\left(\phi_{\tau,K}(\theta, m), \mathcal{D}_\tau^\text{v}\right)\right]\\
        &~\text{s.t.}~ \,\,\; \phi_{\tau,k+1} = \phi_{\tau,k} - \alpha \, \mathbbm{1}_{m\geq 0} \circ \nabla_\phi \mathcal{L}\!\left( \phi_{\tau,k}, \mathcal{D}_\tau^\text{t} \right) ~ \text{and} ~ \phi_{\tau,0}=\theta,
    \end{split}
\end{equation}
This corresponds to setting the outer- and inner-loop loss functions introduced in Section~\ref{sec:MAML-to-sp-MAML} to $\Lo_\tau(\phi)=\mathcal{L}(\phi, \mathcal{D}_\tau^\text{v})$ and  $\Li_\tau(\phi)=\mathcal{L}(\phi, \mathcal{D}_\tau^\text{t})$.

\begin{figure}[b]
\centering
\hspace{-8pt}
\begin{minipage}{.35\textwidth}
  \centering
  \begin{center}
    \includegraphics[width=1\textwidth]{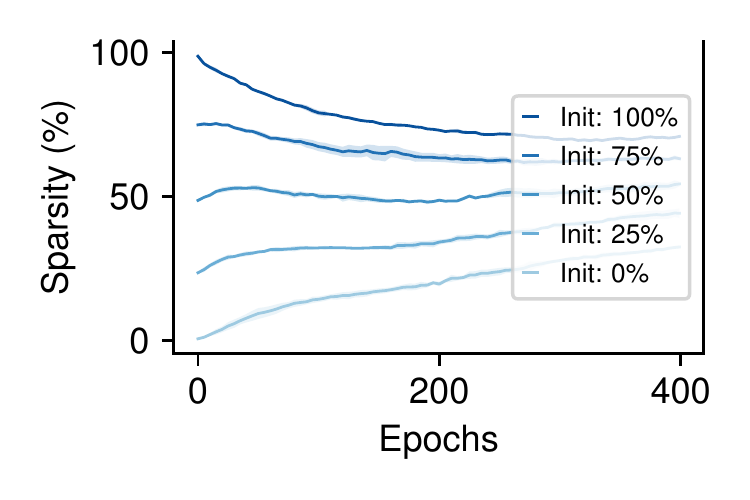}
  \end{center}
  %\vspace{-15pt}
\end{minipage}
\hspace{-12pt}
\begin{minipage}{.35\textwidth}
  \centering
  \begin{center}
    \includegraphics[width=1\textwidth]{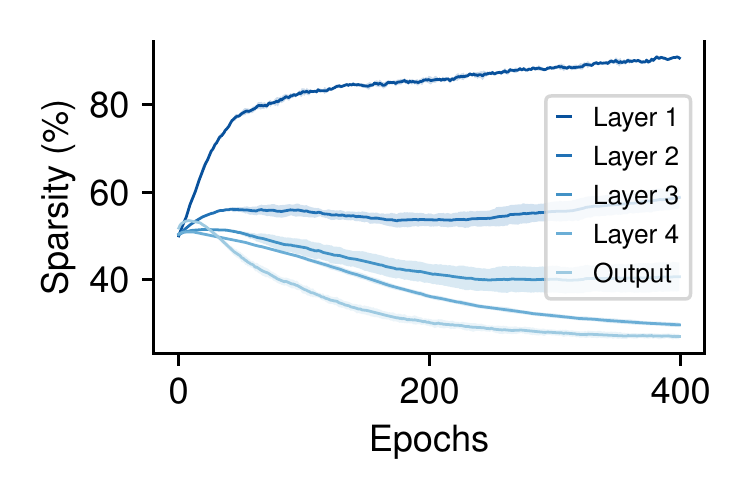}
  \end{center}
  %\vspace{-15pt}
\end{minipage}
\hspace{-12pt}
\begin{minipage}{.35\textwidth}
  \centering
  \begin{center}
    \includegraphics[width=1\textwidth]{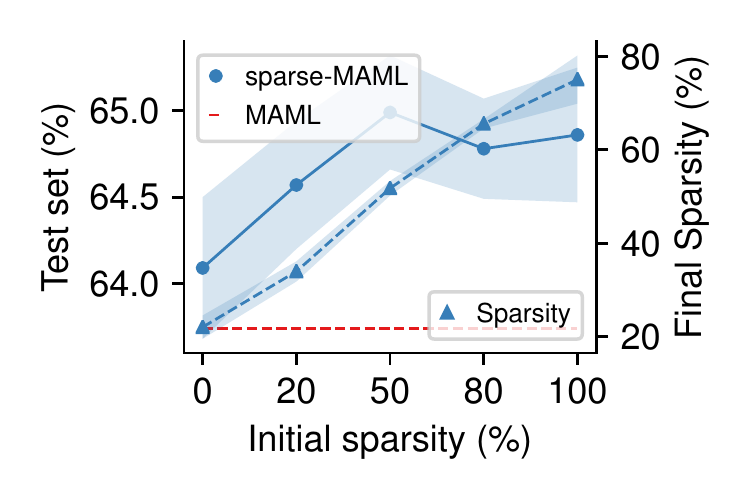}
  \end{center}
  %\vspace{-15pt}
\end{minipage}
  \captionof{figure}{Gradient sparsity emerges in 5-shot, 5-way miniImageNet classification, standard ConvNet model. Results averaged over 5 seeds $\pm$ std. \textit{Left}: Averaged gradient sparsity adapts for different sparsity initializations. \textit{Center}: Different levels of gradient sparsity for convolutional and output layer weights emerge, with gradually less sparsity from earlier to deeper layers, while all being initialized at $\sim50\%$ sparsity. \textit{Right:} Sparse-MAML reaches higher test set accuracy for higher initial levels of gradient sparsity.}
\label{fig:inner_layer_sparse}
%\vspace{-8pt}
\end{figure}

We apply sparse-MAML to the standard few-shot learning benchmark based on the miniImageNet dataset \citep{ravi_2016}. Our main purpose is to understand whether our meta-learning algorithm gives rise to sparse learning by shutting off weight updates, and if the resulting sparse learners achieve better generalization performance. Furthermore, we analyze the patterns of sparsity discovered by sparse-MAML over a range of hyperparameter settings governing the meta-learning process.

Our experimental setup\footnote{Source code available at: \url{https://github.com/Johswald/learning_where_to_learn}} follows refs.~\citep{finn_model-agnostic_2017, vinyals_matching_2017} unless stated otherwise. In particular, by default, our experimental results are obtained using the standard 4-convolutional-layer neural network (ConvNet) model that has been intensively used to benchmark meta-learning algorithms. As is also conventional, we consider two data regimes: 5-shot 5-way, and 1-shot 5-way (the term `shot' denotes the number of examples per class, and `way' the number of classes). As we vary the hyperparameters of our algorithms, we monitor few-shot learning performance and the gradient sparsity level, defined for a parameter group or the entire network as $\lVert \mathbbm{1}_{m< 0} \rVert^2/\text{dim}(m)$. All experimental details can be found in the SM.

\subsection{Gradient sparsity decreases with layer depth}

\begin{wraptable}[22]{R}{0.55\textwidth}
\vspace{-0.4cm}
\centering
  \caption{5-way few-shot classification accuracy (\%) on miniImageNet, standard ConvNet model. We report mean $\pm$ std.~over 5 seeds. All results except ours taken from the respective papers (we use the symbol '---' to indicate missing results). The results for meta-curvature (MC) are not directly comparable as additional data augmentation was used.}
  \label{Tab:results_table}
   \begin{tabular}{lll}
    \toprule
    Method    & 1-shot      & 5-shot\\
    \midrule
    MAML \citep[][]{finn_model-agnostic_2017} & 48.07$^{\pm 1.75}$     & 63.15$^{\pm0.91}$ \\
    ANIL \citep{raghu_rapid_2020} & 46.70$^{\pm 0.40}$     & 61.50$^{\pm0.50}$\\
    BOIL \citep[][]{oh2021boil} & 49.61$^{\pm 0.16}$     & 66.45$^{\pm0.37}$ \\
    \midrule
    Meta-SGD \citep[][]{DBLP:journals/corr/LiZCL17}  & 50.47$^{\pm1.87}$     & 64.03$^{\pm0.94}$\\
    MT-net \citep{lee2018gradient} & 51.70$^{\pm1.84}$ & ---\\
    MC (+data aug.) \citep[][]{DBLP:conf/nips/ParkO19}  & 54.23$^{\pm0.88}$     &68.47$^{\pm0.69}$\\
    Shrinkage \citep{DBLP:journals/corr/abs-1909-05557} & 47.7$^{\pm0.5}$     &---\\
    \midrule
    exp-MAML &48.38$^{\pm0.45}$ & 65.21$^{\pm0.62}$ \\
    sparse-ReLU-MAML &50.39$^{\pm0.89}$ &64.84$^{\pm0.46}$ \\
    sparse-MAML &50.35$^{\pm0.39}$ &67.03$^{\pm0.74}$ \\
    sparse-MAML$^+$ &51.04$^{\pm 0.59}$ &68.05$^{\pm0.84}$ \\
    \bottomrule
  \end{tabular}
 %\vspace{-8pt}
\end{wraptable}

Our first finding validates and extends the phenomena described by Raghu et al.~\citep{raghu_rapid_2020} and Chen et al.~\citep[][]{DBLP:journals/corr/abs-1909-05557}. As shown in Figure~\ref{fig:inner_layer_sparse},
sparse-MAML dynamically adjusts gradient sparsity across the network, with very different values over the layers. As an example, we show the average gradient sparsity of the four convolutional weight matrices and the output layer during training. The same trend is observed for other parameter groups in the network except the output bias (for which sparsity is always high; see SM). Sparsity clearly correlates with depth and gradually increases towards the early layers of the network, despite the similar value before training (around 50\%), i.e., sparse-MAML suppresses inner-loop updates of weights in earlier layers while allowing deeper layers to adjust to new tasks. This effect is robust across different sparsity initializations, with final few-shot learning performance correlating with sparsity, cf.~Figure~\ref{fig:inner_layer_sparse}.

These findings validate that our method can discover sparse learning algorithms. Moreover, they show that the level of sparsity is anti-correlated with depth. This result can be interpreted in the light of neural network models with human-engineered patterns of frozen features, which freeze layers of features based on their depth (in combination with MAML, see e.g.~ANIL and BOIL, \citep{raghu_rapid_2020,oh2021boil}). Our method justifies these approaches, while outperforming them, cf.~Table~\ref{Tab:results_table}, suggesting that it might be preferable to meta-learn which features to freeze. We note that another related method for automatic discovery of task-shared weights based on learning per-parameter $L_2$ regularization strengths \citep[Shrinkage,][]{DBLP:journals/corr/abs-1909-05557} yields a similar trend of high freezing for lower-level features, without however improving performance against standard MAML. Our findings hold when applying our method to a deeper and wider residual neural network (ResNet-12) model, see Tables~\ref{Tab:resnet_results_table}~and~\ref{Tab:all_resnet_12_results} (SM), where we observe the same trend of decreasing gradient sparsity with depth emerge.

\begin{figure}[h!]
\centering
\hspace{-10pt}
\begin{minipage}{.50\textwidth}
  \centering
  \begin{center}
    \includegraphics[width=0.85\textwidth]{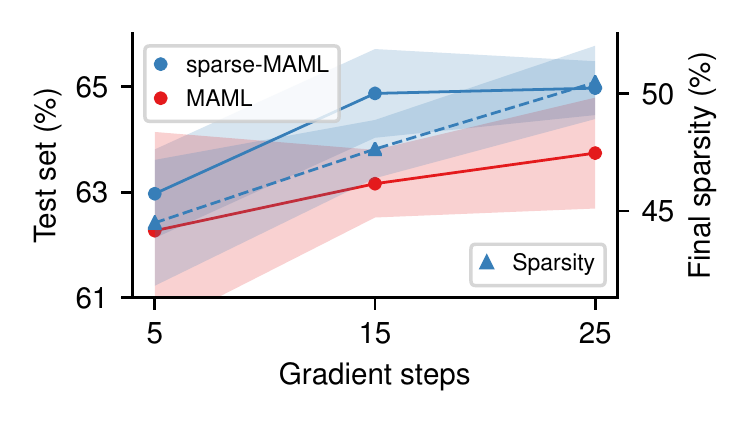}
  \end{center}
  %\vspace{-15pt}
\end{minipage}
\hspace{-12pt}
\begin{minipage}{.5\textwidth}
  \centering
  \begin{center}
    \includegraphics[width=0.9\textwidth]{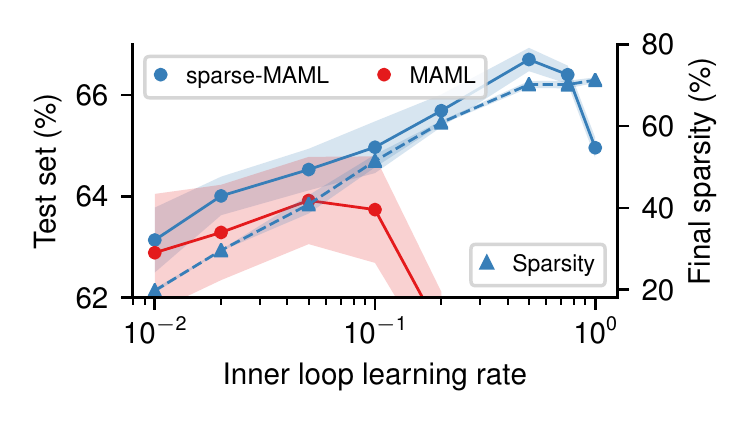}
  \end{center}
  \vspace{-15pt}
\end{minipage}
  \captionof{figure}{Sparse learning algorithms found by sparse-MAML work best in combination with highly-plastic models. Both gradient sparsity and generalization performance increase with number of inner-loop learning steps (\textit{left}) and learning rate (\textit{right}). Standard MAML, which does not employ sparse learning, requires more careful learning rate tuning and does not benefit as much from large learning rates. In all experiments, gradient sparsity is initially $\sim50\%$. The inner-loop learning rate is set to $0.1$ when varying the number of steps. Results are for 5-shot, 5-way miniImageNet, averaged over 5 seeds $\pm$ std.
  }
  \label{fig:number_of_steps}
 \vspace{-15pt}
\end{figure}

\subsection{Sparse learning prefers highly-plastic models}

We hypothesize that restricting learning to an appropriate parameter subset allows for longer training and larger changes without overfitting, beyond meta-learning initial parameter values. To verify this hypothesis we scan over different inner-loop learning rates and lengths and compare the resulting test set performances of MAML and sparse-MAML. 

First, we test three different inner-loop durations (5, 15 or 25 gradient steps, see Figure~\ref{fig:number_of_steps}, left). We find that neither MAML nor sparse-MAML exhibit overfitting for the duration range considered here (for reference, the original study of MAML applied 5 inner-loop steps during meta-training). In contrast to MAML, the solutions found by sparse-MAML generalize significantly better for longer adaptation phases. This improvement in generalization performance is accompanied by an increase in gradient sparsity. Furthermore, applying sparse-MAML in the very-low data regime of 1-shot learning results in higher levels of gradient sparsity, even though the exact same model and training setup is used for both 1- and 5-shot learning experiments.

We further investigate if increasing the learning rate can result in improved generalization performance in combination with sparse learning. We scan the inner-loop learning rate over a large range, cf.~Figure~\ref{fig:number_of_steps} (right), and find a clear trend towards gradient sparsity going along with better test-set accuracy for larger learning rates. Interestingly, similar effects have been reported in standard (non-meta-learned) neural network training where both freezing layers throughout training \citep{raghu_svcca_2017,brock_freezeout_2017} and the use of large learning rates \citep{lewkowycz_large_2020} seem to improve generalization performance.

\subsection{Sparse learning vs.~more expressive gradient modulation methods}

\begin{wraptable}[17]{R}{0.55\textwidth}
\vspace{-0.40cm}
\centering
  \caption{5-way few-shot classification accuracy (\%) on miniImageNet with a ResNet-12 model. We report mean $\pm$ std.~over 3 seeds. We report MetaOptNet \cite{meta_opt_net} figures when no additional regularization techniques are applied. Results from BOIL and MetaOptNet are taken from the respective papers.}
  \label{Tab:resnet_results_table}
   \begin{tabular}{lll}
    \toprule
    Method    & 1-shot      & 5-shot\\
        \midrule
    MetaOptNet & 51.13  &  70.88 \\
    \midrule
    MAML  & 53.91$^{\pm 0.61}$     & 69.36$^{\pm1.23}$ \\
    ANIL & 55.25$^{\pm 0.33}$     & 70.03$^{\pm0.58}$\\
    BOIL & ---     & 70.50 $^{\pm0.28}$\\
    \midrule
    sparse-MAML & 55.02$^{\pm0.46}$ & 70.02$^{\pm1.12}$ \\
    sparse-ReLU-MAML & 56.39$^{\pm0.38}$ & 73.01$^{\pm0.24}$ \\
    \bottomrule
  \end{tabular}
 %\vspace{-8pt}
\end{wraptable}
Sparse-MAML can be understood as a binary gradient modulation method. Second-order methods such as meta-curvature \citep[][]{DBLP:conf/nips/ParkO19} modulate gradients by meta-learning pre-conditioning matrices; in meta-SGD \citep{DBLP:journals/corr/LiZCL17}, these matrices are restricted to be diagonal; sparse-MAML further restricts the diagonal values to be binary. From this point of view, sparse-MAML is the least expressive form of gradient modulation.

\begin{wraptable}[10]{R}{0.55\textwidth}
\vspace{-0.4cm}
\centering
  \caption{Average gradient sparsity levels (\%) after meta-learning on 5-way miniImageNet few-shot tasks, standard ConvNet model.  Mean $\pm$ std.~over 5 seeds.}
  \label{Tab:sparsity_table}
   \begin{tabular}{lll}
    \toprule
    Method   & 1-shot      & 5-shot \\
   
    \midrule
    sparse-ReLU-MAML &72.62$^{\pm0.73}$ &70.92$^{\pm0.85}$ \\
    sparse-MAML &79.04$^{\pm1.61}$ &74.98$^{\pm0.10}$ \\
    sparse-MAML$^+$ &78.05$^{\pm 1.67}$ &76.66$^{\pm1.13}$ \\
    \bottomrule
  \end{tabular}
\end{wraptable}
Surprisingly, we find that despite its reduced expressiveness, sparse-MAML recovers the performance improvements achieved by the more sophisticated alternatives, significantly improving the performance of standard MAML (cf.~Table \ref{Tab:results_table}).
We point out that sparse-MAML uses a first-order update (Eq.~\ref{eq:sp-MAML-mask-update}), while all three gradient modulation methods we compare to (meta-SGD, meta-curvature and MT-nets) use second-order derivatives that are more costly to evaluate.

\paragraph{Sparse learning emerges when meta-learning learning rates.} We also implement a variant of meta-SGD which uses rectified learning rates (sparse-ReLU-MAML).  Concretely, we replace the step function $\mathbbm{1}_{m\geq 0}$ in the inner-loop dynamics (Eq.~\ref{eq:sp-MAML-inner-loop}) by the positive part of $m$, $(m)_+ := \mathbbm{1}_{m\geq 0} \, m$. Then, we learn the underlying learning rate parameter $m$ using our first-order straight-through update of Eq.~\ref{eq:sp-MAML-mask-update} to prevent learning rates from getting stuck at zero. Besides standard meta-SGD, which allows learning rates to go negative, we compare this method to an alternative exponential learning rate parameterization \citep{sutton1992adapting}, $\exp m$, which like sparse-ReLU-MAML enforces non-negativity while avoiding permanently frozen updates (exp-MAML, Table~\ref{Tab:results_table}). It is, however, harder to reach sparse learning rate distributions under this parameterization, as the meta-gradient $\mathrm{d}_m \, \mathcal{L}$ becomes exponentially small as $m$ approaches zero.

We analyze the distributions of learning rates that sparse-ReLU-MAML yields on miniImageNet and observe that gradient sparsity once more emerges, cf.~Table~\ref{Tab:sparsity_table}. We find that the levels of gradient sparsity when meta-learning binary (sparse-MAML) or rectified learning rates (sparse-ReLU-MAML) are approximately the same, with sparse-MAML performing better than sparse-ReLU-MAML on 5-shot tasks, and better than exp-MAML on both 1-shot and 5-shot tasks. These results support the hypothesis that shutting off weight updates is one of the essential gradient modulation operations in few-shot learning. We note that while sparse-MAML and sparse-ReLU-MAML quickly disable learning in a large fraction of weights, exp-MAML tends to push learning rates down, in particular for layers close to the input, but at a much slower pace; increasing the meta-learning rate $\gamma_m$ cannot compensate for this slowdown as learning becomes unstable (data not shown).

This picture changes for the the deeper and larger ResNet-12 model, cf.~Table~\ref{Tab:resnet_results_table}.  When using this more complex architecture, we find that sparse rectified learning rates (sparse-ReLU-MAML) are beneficial over binary gradient masks (sparse-MAML). In particular, the combination of sparse learning (see Table~\ref{Tab:all_resnet_12_results} for additional details) with learning rate modulation found by sparse-ReLU-MAML outperforms all other methods, including standard (dense-learning) MAML, as well as methods based on manually freezing layers in the inner-loop: BOIL \citep[][]{oh2021boil}, ANIL \citep{raghu_rapid_2020}, and the closely-related MetaOptNet \citep[][]{meta_opt_net} method. Like ANIL, MetaOptNet only adapts the final classification layer in the inner-loop, but it uses a more sophisticated solver instead of a few steps of gradient descent to learn task-specific solvers. Thus, once more, learning by sparse gradient descent is an effective strategy to improve the generalization performance of a few-shot learner.

\begin{table}[b]
\vspace{-0.6cm}
\centering
  \caption{Few-shot classification accuracy (\%) when meta-learning on miniImageNet but meta-testing on TieredImageNet, CUB and Cars. Mean $\pm$ std.~over 5 seeds.} 
  \label{Tab:boil}
   \begin{tabular}{lllll}
    \toprule
    Problem & Method  & TieredImageNet & CUB & Cars\\
    \midrule
    
    \multirow{6}{*}{1-shot} & MAML  & 51.61$^{\pm 0.20}$     & 40.51$^{\pm 0.08    }$ & 33.57$^{\pm 0.14}$  \\ 
    & ANIL & 52.82$^{\pm 0.29}$     & 41.12$^{\pm 0.15}$ & 34.77$^{\pm 0.31}$  \\ 
    & BOIL  & 53.23$^{\pm 0.41}$     & \textbf{44.20$^{\pm 0.15}$} & 36.12$^{\pm 0.29}$  \\
    \cmidrule(lr{1em}){2-5}
    & sparse-ReLU-MAML  & 53.77$^{\pm 0.94}$     & 42.89$^{\pm 0.45}$ & 36.04$^{\pm 0.55}$  \\ 
    & sparse-MAML   & 53.47$^{\pm 0.53}$     & 41.37$^{\pm 0.73}$ & 35.90$^{\pm 0.50}$ \\ 
    & sparse-MAML$^+$  & \textbf{53.91$^{\pm 0.67}$}     & 43.43$^{\pm 1.04}$ & \textbf{37.14$^{\pm 0.77}$}\\

    \midrule
    \multirow{6}{*}{5-shot} & MAML  & 65.76$^{\pm 0.27}$     & 53.09$^{\pm 0.16}$ & 44.56$^{\pm 0.21}$  \\ 
    & ANIL   & 66.52$^{\pm 0.28}$     & 55.82$^{\pm 0.21}$ & 46.55$^{\pm 0.29}$  \\  
    &BOIL   & 69.37$^{\pm 0.23}$     & 60.92$^{\pm 0.11}$ & 50.64$^{\pm 0.22}$  \\
    \cmidrule(lr{1em}){2-5}
    & sparse-ReLU-MAML  & 68.12$^{\pm 0.69}$     & 57.53$^{\pm 0.94}$ & 49.95$^{\pm 0.42}$  \\ 
    & sparse-MAML   & 68.83$^{\pm 0.65}$     & 60.58$^{\pm 1.10}$ & 52.63$^{\pm 0.56}$  \\ 
    & sparse-MAML$^+$ & \textbf{69.92$^{\pm 0.21}$}     & \textbf{62.02$^{\pm 0.78}$} & \textbf{53.18$^{\pm 0.44}$}\\ 
    \bottomrule
  \end{tabular}
\end{table}

\paragraph{Stochastic gradient masking.} We further investigate whether stochastic binary gradient masks can improve few-shot learning performance. Our interest in studying stochastic masks is two-fold: as a way to improve meta-optimization based on our straight-through estimator; and to determine if stochastic masking is beneficial at meta-test time. We thus investigate sparse-MAML$^+$, a variant of our algorithm in which gradient masks are generated from a low-dimensional Gaussian vector, with noise intensity determined by meta-learning (see SM).  As before, we adjust meta-parameters using a first-order update. We find that this mask generation method does result in improved performance, cf.~Table~\ref{Tab:results_table}. Interestingly, mask randomness is entirely suppressed by meta-learning; eventually, $\sigma \to 0$, and we recover a single deterministic mask $m$. The performance improvements observed on few-shot learning therefore stem from improvements to the meta-optimization process, likely related to the challenges of optimizing binary variables with (pseudo)gradient-based methods.

\subsection{Sparse learning improves performance in cross-domain adaptation tasks}
We now investigate whether the patterns of gradient sparsity discovered by our method overfit to the particular task family where they were obtained, namely, to few-shot miniImageNet classification tasks. This is an important question, since excessive parameter freezing may prevent adaptation to tasks that are too different from those presented during meta-learning.

We therefore move our analysis of few-shot learning to a cross-domain adaptation setting. In cross-domain adaptation problems, the family of tasks presented post-meta-learning to evaluate our algorithms is shifted by sampling classes from a different dataset.
In particular, we train our meta-learner on the miniImageNet dataset and then evaluate learning performance on the TieredImageNet, CUB and Cars datasets.
It has previously been demonstrated that manually freezing either the head (BOIL) or the body (ANIL) during meta-testing improves performance in this setting \citep{oh2021boil}, compared to letting all weights adapt (MAML).
In Table~\ref{Tab:boil} we compare the performance of our method to these baselines. We find that meta-learning the freezing pattern with sparse-MAML as opposed to manually selecting it consistently improves cross-domain adaptation.

\section{Continual learning}

We now turn to a continual learning setting, where tasks must be learned sequentially. A successful continual learner is able to learn similar tasks faster, as in the few-shot learning case, while retaining high performance on previously seen tasks. We conjecture that sparse learning can improve memory retention and accelerate future learning by reducing interference with past updates.

\subsection{Gradient sparsity emerges when learning continually with Look-ahead MAML}

We investigate the benefits of sparse gradients in the recently proposed La-MAML algorithm \citep{gupta_-maml_2020}.  This algorithm combines online meta-learning in conjunction with a small replay buffer which holds representative examples from the past in memory. Standard replay methods \citep{robins1995catastrophic}, define a joint objective using present and buffered data and directly optimize this objective. La-MAML follows a technique known as meta-experience replay \citep{riemer2019learning} and introduces a bi-level optimization problem. The outer loss $\Lo$ is the multi-task objective optimized with standard replay methods, while the inner loss $\Li$ is evaluated on the new incoming data only. \citet{riemer2019learning} have shown that such meta-learning promotes gradient alignment over tasks, which is a way to reduce interference \citep{lopez2017gradient}.

\begin{figure}[b]
\centering
\hspace{-10pt}
\begin{minipage}{.35\textwidth}
  \centering
  \begin{center}
    \includegraphics[width=1\textwidth]{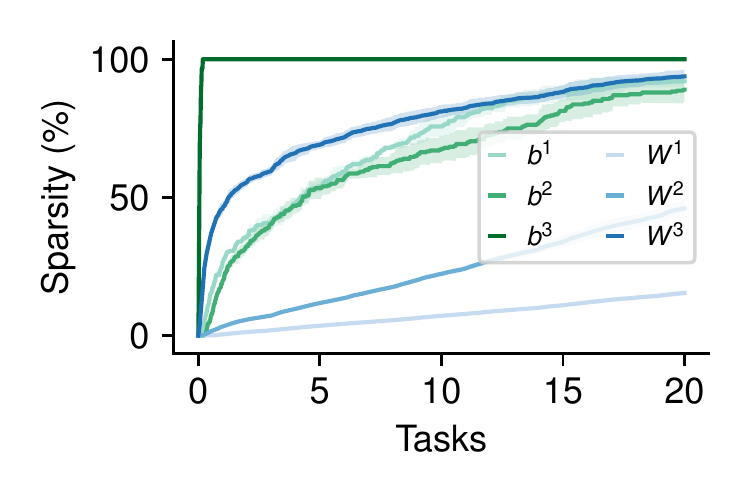}
  \end{center}
  %\vspace{-15pt}
\end{minipage}
\hspace{-12pt}
\begin{minipage}{.35\textwidth}
  \centering
  \begin{center}
    \includegraphics[width=1\textwidth]{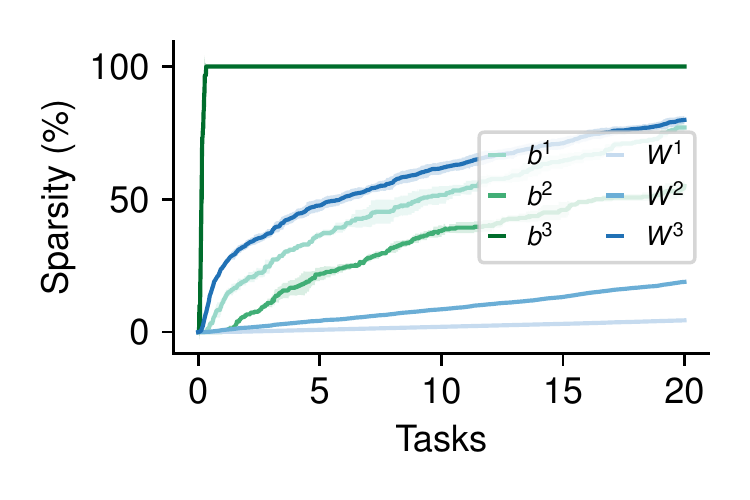}
  \end{center}
  %\vspace{-15pt}
\end{minipage}
\hspace{-12pt}
\begin{minipage}{.35\textwidth}
  \centering
  \begin{center}
    \includegraphics[width=1\textwidth]{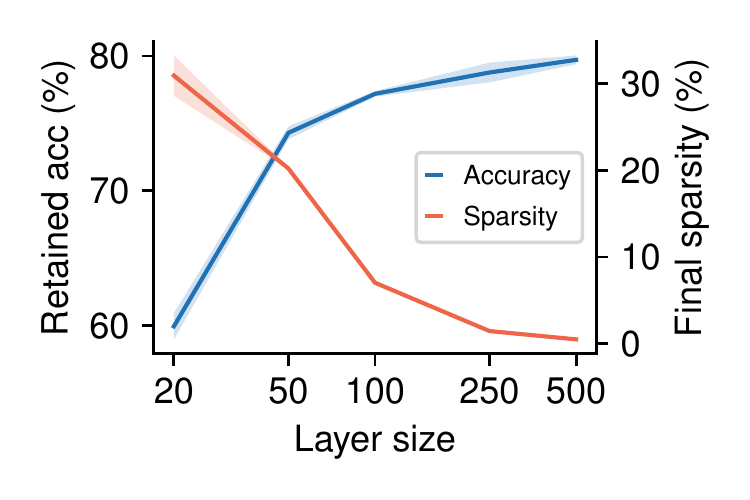}
  \end{center}
  %\vspace{-15pt}
\end{minipage}
  \captionof{figure}{Gradient sparsity when learning MNIST rotations with the La-MAML and sparse-La-MAML algorithms. Results averaged over 3 seeds $\pm$ std. \textit{Left}: Sparsity emerges on the original La-MAML algorithm across the three layer network and monotonically increases with the number of tasks and with depth for both weight ($W_1, W_2, W_3$) and bias parameters ($b_1, b_2, b_3$). \textit{Center}: A similar behavior is observed when replacing meta-learned learning rates by meta-learned binary gradient masks (sparse-La-MAML).  \textit{Right:} Overall sparsity of sparse-La-MAML  decreases with increased network capacity accompanied with higher retained accuracy (RA). Network capacity is varied by changing the number of neurons in the two hidden layers simultaneously.}
\label{fig:lamaml_sparsity}
%\vspace{-8pt}
\end{figure}

\begin{wrapfigure}[16]{R}{0.4\textwidth}
\vspace{-0.7cm}
  \begin{center}
    \includegraphics[width=0.4\textwidth]{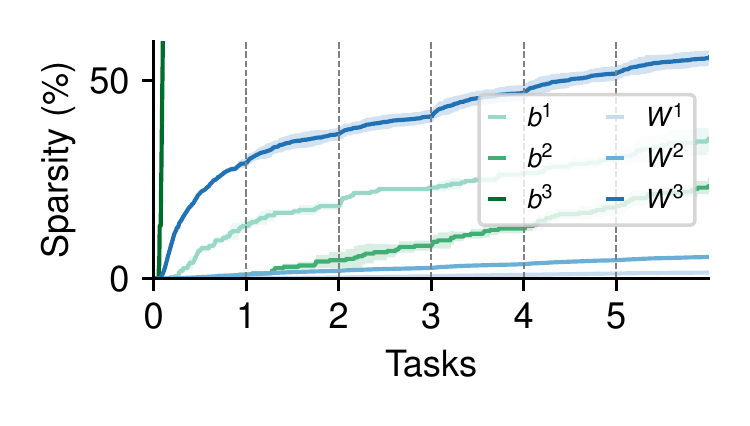}
  \end{center}
  \vspace{-0.5cm}
  %\vspace{-15pt}
  \captionof{figure}{Structured sparsity emerges and tends to converge within a task in multi-pass continual learning. Results shown for La-MAML after training on MNIST rotations, averaged over 3 seeds $\pm$ std., for weight layers ($W_1, W_2, W_3$) and bias parameters ($b_1, b_2, b_3$). Sparsity increases with depth.}
  \label{fig:la-maml-multipass-sparsity}
\end{wrapfigure}

Like the variants of MAML reviewed in Section~\ref{sec:MAML-to-sp-MAML}, La-MAML introduces meta-learned per-parameter learning rates. We now briefly review a single iteration of the algorithm; complete pseudocode is provided in the SM. Each iteration of La-MAML consists of processing a new batch of data $\mathcal{B}$ as follows: (\emph{i}) starting from $\phi_0 = \theta$ taking an inner-loop step on each sample $k$ in $\mathcal{B}$, with $\Li_k(\phi_k) = \mathcal{L}(\phi_k, \mathcal{B}_k)$; (\emph{ii}) defining an outer-loss $\Lo(\phi_K, \mathcal{B} \cup \mathcal{R}$) on both new data $\mathcal{B}$ and a batch of past data $\mathcal{R}$ sampled from the replay buffer; (\emph{iii}) taking an outer-loop step on the learning rate parameter using a first-order update followed by an outer-loop step (using the newly updated learning rate) on the neural network parameters $\theta$; (\emph{iv}) re-populating the replay buffer with data in $\mathcal{B}$. The sequence of inner-loop updates is given by
\begin{equation}
    \label{eq:la-maml-inner}
    \phi_{k+1} = \phi_{k} - (\alpha)_+ \circ \nabla_\phi \Li_k(\phi_k),\qquad\text{s.t. }\phi_0 = \theta,
\end{equation}
where $\alpha$ is a vector of learning rates whose components are constrained to be non-negative by elementwise application of the positive part function. We note that while the main text of ref.~\cite{gupta_-maml_2020} presents an inner-loop learning rate parameter that is allowed to go negative, the implementation for the experiments reported in ref.~\citep[][]{lamamlcode} uses rectified learning rates. In this implementation, a learning rate that is updated below zero will never recover, which can lead to dead coordinates and promote sparsity. The inner loss $\Li_k(\phi)$ is defined on a different data sample on each step $k$. A first-order update is applied to $\theta$, again modulated by the adaptive learning rate:
\begin{equation}
    \label{eq:la-maml-outer}
    \theta \gets \theta - (\alpha)_+\circ \nabla_\phi \Lo(\phi_K).
\end{equation}

\paragraph{Sparse-La-MAML.} Our sparse-MAML can be readily applied to continual learning problems by modifying the inner- and outer-loop updates of La-MAML. We replace the meta-learned learning rates in equations~\ref{eq:la-maml-inner}-\ref{eq:la-maml-outer} by meta-learned binary gradient masks, $\alpha = \alpha_0 \, \mathbbm{1}_{m_\geq 0}$, with $\alpha_0 \in \mathbb{R}_+$ some scalar (fixed) learning rate value.
To learn the underlying parameter $m$, we again resort to our first-order update (equation~\ref{eq:sp-MAML-mask-update}).

\begin{comment}
We refer to the resulting algorithm summarized below as sparse-La-MAML:
\begin{align}
    \phi_{k+1} &= \phi_{k} - \alpha_0 \, \mathbbm{1}_{m_\geq 0} \circ \nabla_\phi \Li_k(\phi_k),\quad \text{for } k=1,\ldots,K, \quad\text{s.t. }\phi_0 = \theta,\label{eq:sp-lam-phi}\\
    \theta &\gets \theta - \alpha_0 \, \mathbbm{1}_{m_\geq 0} \circ \nabla_\phi \Lo(\phi_K)\label{eq:sp-lam-theta}\\
    m &\gets m + \alpha_0 \, \gamma_m \,  \nabla_\phi \Lo\!\left(\phi_{K}\right) \circ \sum_{k=0}^{K-1} \nabla_\phi\Li_k\!\left( \phi_{k}\right).\label{eq:sp-lam-m}
\end{align}
\end{comment}

\paragraph{Sparse learning improves continual learning.} We hypothesize that a large fraction of learning rates approaches zero when the hyperparameters of La-MAML and sparse-La-MAML are tuned for best continual learning performance. To test this hypothesis, we follow the exact same setup as in the original study of La-MAML  \cite{gupta_-maml_2020}. We perform a grid search over the learning rates $\alpha_0$ and $\gamma_m$, and search for best continual learning performance, not sparsity (cf.~SM). The remaining hyperparameters are kept to the values provided in \cite{gupta_-maml_2020}.

We study the three MNIST \citep{lecun_mnist_1998} continual learning problems \textit{rotations}, \textit{permutations} and \textit{many permutations} using a single-headed network using the code accompanying ref.~\citep{gupta_-maml_2020}. Task information is not given to the network, and each data point is seen only once, unless noted otherwise. Full details as well as additional experiments using the CIFAR-10 \citep[][]{krizhevsky_learning_2009} dataset are provided in the SM.

We verify that our initial hypothesis is correct: La-MAML shuts off learning in many coordinates (cf.~Figure~\ref{fig:lamaml_sparsity}; full results may be found in the SM, Table~\ref{Tab:sparsity_values}), reaching even higher levels of sparsity than sparse-La-MAML. This can be explained by the fact that dead coordinates can arise in La-MAML, which can lead to excess sparsity. By contrast, our straight-through update dynamically and continually adjusts the pattern of sparsity allowing previously frozen parameters to be unfrozen. This results in matching or slightly improved performances when using our binary gradient mask across all three MNIST variants, see Table~\ref{Tab:la_maml_mnist}, both in terms of final retained accuracy (RA) and backward-transfer and interference (BTI; the change in accuracy measured at the end of the experiment minus just after learning a task, averaged over tasks). Moreover, the patterns of sparsity adjust to the capacity of the network, decreasing and eventually vanishing for larger models (Figure~\ref{fig:lamaml_sparsity}), as retained accuracy goes up, indicating that the task is not sufficiently difficult to create interference on large capacity models.

\begin{table}[h!]
\vspace{-0.2cm}
\centering
  \caption{Retained accuracy (RA) and backward-transfer and interference (BTI) for three different MNIST continual learning problems: rotations, permutations and many permutations. We report mean $\pm$ std.~over 5 seeds. Negative BTI values closer to zero imply less forgetting and are therefore better. Results of related work are taken from \citep{gupta_-maml_2020}; for completeness we include the GEM \citep[][]{lopez2017gradient} and MER \citep{riemer2019learning} methods next to a stochastic gradient descent baseline. Although sparse-La-MAML (sp-LaM) is strictly less expressive than the original La-MAML algorithm, it shows competitive performance across all variants and both metrics. The lower baseline BTI values can be explained by lower overall accuracies achieved by La-MAML.}
  \label{Tab:la_maml_mnist}
   \begin{tabular}{lllllll}
    \toprule
    Method  & \multicolumn{2}{c}{Rotations} & \multicolumn{2}{c}{Permutations} &
    \multicolumn{2}{c}{Many permutations}\\
    \midrule
        & RA       & BTI  & RA       & BTI & RA       & BTI \\
    \midrule
    Baseline & 53.38$^{\pm 1.53}$     & \textbf{-5.44$^{\pm 1.70}$} & 55.42$^{\pm 0.65}$     & -13.76$^{\pm 1.19}$ & 32.62$^{\pm 0.43}$     & -19.06$^{\pm  0.86}$\\
    GEM & 67.38$^{\pm 1.75}$     & -18.02$^{\pm 1.99}$ & 55.42$^{\pm 1.10}$     & -24.42$^{\pm 1.10}$ & 32.14$^{\pm 0.50}$     & -23.52$^{\pm  0.87}$\\
    MER  & 77.42$^{\pm 0.78}$     & -5.60$^{\pm 0.70}$ & 73.46$^{\pm  0.45}$     & -9.96$^{\pm 0.45}$ & 47.40$^{\pm 0.35}$     & -17.78$^{\pm  0.39}$\\
    \midrule
    La-M  & 77.42$^{\pm 0.65}$     & -8.64 $^{\pm 0.40}$ & 74.34$^{\pm  0.67}$     & \textbf{-7.60$^{\pm 0.51}$} & 48.46$^{\pm 0.45}$     & \textbf{-12.96$^{\pm  0.07}$}\\
    sp-LaM  & \textbf{77.77$^{\pm 0.58}$}     & -8.16 $^{\pm 0.61}$ & \textbf{76.88$^{\pm  0.72}$}     & -8.39$^{\pm  0.63}$ & \textbf{50.81$^{\pm  0.79}$}     & -13.73$^{\pm  0.73}$ \\
    \bottomrule
  \end{tabular}
\end{table}

\begin{wraptable}[19]{R}{0.55\textwidth}
\vspace{-0.4cm}
\centering
  \caption{Sparse learning improves continual-MAML performance. Cumulative online accuracy on  Omniglot-MNIST-FashionMNIST benchmark. Tasks switch with probability $1-p$. Results from previous work taken from \citep{caccia2020online}. Mean $\pm$ std.~over 5 seeds.}
  \label{Tab:osaka}
   \begin{tabular}{lll}
    \toprule
    Method    &  $p=0.98$  &  $p=0.9$ \\
    \midrule
    Online Adam \citep{kingma_adam_2015} & 73.9\std{2.2}  & 23.8\std{1.2}  \\
        Fine-tuning & 72.7\std{1.7}  & 22.1 \std{1.1} \\
        MAML \citep{finn_model-agnostic_2017} & 84.5\std{1.7} &  75.5\std{0.7} \\
        ANIL \citep{raghu_rapid_2020} & 75.3\std{2.0} &   69.1\std{0.8}  \\
        BGD \citep{DBLP:journals/corr/abs-2010-00373} & 87.8\std{1.3}  & 63.4\std{0.9} \\
        MetaCOG  \citep{he2019task} & 88.0\std{1.0} &   63.6\std{0.9} \\
        MetaBGD  \citep{he2019task} & 91.1\std{2.6} &   74.8\std{1.1} \\
       \midrule
        C-MAML  &  92.8\std{0.6}  & 83.3\std{0.4} \\
        sparse-C-MAML &  94.2$^{\pm0.4}$ & 86.3\std{0.4}\\
        sparse-ReLU-C-MAML & 93.5\std{0.5} & 86.1\std{0.2} \\ 
    \bottomrule
  \end{tabular}
\end{wraptable}
As in our few-shot learning experiments, structured sparsity emerges across the different parameter groups of the network (cf.~Figure~\ref{fig:lamaml_sparsity}). We observe that now sparsity is highest closest to the output layer, the exact opposite of the trend found in our few-shot learning experiments. This provides evidence that online meta-learning can discover how to rewire low-level features without interference in order to accommodate different tasks that share high-level structure. 
We further investigate a multi-pass setting, where the examples from each task are visited multiple times (10 epochs instead of 1) before proceeding to the next task. In this setting, it can be seen that sparsity levels (displayed in Figure~\ref{fig:la-maml-multipass-sparsity}) tend to converge within tasks and then raise again when tasks switch, presumably to preserve past memories via gradient sparsification.  Taken together, our results support the hypothesis that gradient sparsity is beneficial for continual learning and that appropriate patterns of sparsity can be discovered by simple online gradient-based meta-learning.

\subsection{Sparse online learning}

We finally consider another online learning setting in which the underlying task is concealed from the learner and can randomly change at each step, potentially going back to previously seen tasks \citep{ritter2018been,he2019task,caccia2020online}. At each time step $t$, the data $\mathcal{D}_t$ is an i.i.d.~sample from a stationary distribution that only depends on the current task. The learner, whose current state is denoted by $\phi_t$, is evaluated whenever new data is presented and modifies its behavior accordingly. The goal is then to minimize the cumulative loss $\sum_{t=1}^T \mathcal{L}(\phi_t, D_t)$ measuring the performance of the learner before adaptation takes place.

This online learning protocol differs from the one adopted in the previous section, where tasks were visited only once and only the final loss $\sum_{t=1}^T \mathcal{L}(\phi_T, \mathcal{D}_t)$ evaluated at $\phi_T$ mattered. The cumulative loss criterion emphasizes fast learning and adaptability while memory is still needed to avoid re-learning, since tasks can be re-encountered. Recently, it has been shown that a simple modification of MAML \citep[continual-MAML;][]{caccia2020online} can outperform a number of algorithms specifically tailored for this setting as well as plain stochastic gradient descent \citep{Bottou98on-linelearning}. Briefly, continual-MAML extends MAML by introducing a task-switch detection mechanism based on changes in loss; data is buffered until a switch is detected. When this occurs, the buffered data is used to perform a meta-parameter update; the buffer is reset; and the inner-loop optimization restarts. Here, we merge continual-MAML with sparse-MAML, and modulate inner-loop gradients according to \eqref{eq:sp-MAML-inner-loop}. We present complete pseudocode for the algorithm in the SM.

We reproduce the experiments of \cite{caccia2020online} in which a sequence of 10000 examples from the Omniglot \citep[][]{lake_one_2011}, MNIST \citep{lecun_mnist_1998} and FashionMNIST \citep{xiao_fashion-mnist_2017} datasets is presented for online learning to a single-headed neural network, using the code provided by the authors. We carry out a grid search to tune the inner-loop learning rate $\alpha_0$ and the mask learning rate $\gamma_m$ introduced by sparse-MAML for best performance, not sparsity (see SM). We observe again structured (layer-dependent) gradient sparsity emerge when using this algorithm (sparse-C-MAML), as shown in Figure~\ref{fig:sparsity-osaka}. Moreover, gradient sparsity is accompanied by an increase in cumulative online learning accuracy over the original continual-MAML algorithm (cf.~Table~\ref{Tab:osaka}). Finally, we observe the same qualitative behavior (see SM for sparsity levels) and obtain similar performance when replacing our binary masks by rectified learning rates (sparse-ReLU-C-MAML) meta-learned with our straight-through update. These findings once more support the hypothesis that sparse learning, and not learning rate modulation, lead to the improved performance reported here.

\section{Discussion}
We studied gradient-based meta-learning systems with the ability of learning where to learn. This was modeled by adding binary variables which masked gradients on a per-parameter basis, therefore determining which parameters are allowed to change. We observed gradient sparsity emerge in standard few-shot and continual learning problems, without introducing an explicit bias towards sparsity. This form of sparse learning, which may be understood as sparse gradient descent, was accompanied by overall improvements in generalization, as well as reduced interference and forgetting.

Previous work on gradient modulation has focused on estimating task-shared loss geometry to precondition the optimization procedure \citep{DBLP:journals/corr/LiZCL17, pmlr-v97-zintgraf19a, flennerhag2020metalearning, lee2018gradient, zhao_meta_learning_hypernetworks}. In addition, a stochastic variant of gradient masking was featured in the MT-net algorithm \citep{lee2018gradient} as part of a more complex model. Our approach differs from these previous studies in its simplicity. We restrict gradient modulation to be binary and deterministic and use an inexpensive first-order update to learn the gradient masks. In contrast to more traditional methods for inducing sparsity via regularization \citep{tibshirani1996regression} (here, gradient regularization) our approach does not require evaluating second derivatives, which would result from differentiating gradient regularizers. Despite these simplifications, we find competitive performance on our experiments. These results point towards sparse gradient descent as a powerful learning principle.

The idea of meta-learning learning rates can be traced back to the seminal work of Sutton \citep{sutton1992adapting}, who proposed to estimate learning rate meta-gradients online using forward-mode automatic differentiation, and to use consecutive batches of data to define inner- and outer-loop loss functions. This approach, known as stochastic meta-descent (SMD), was extended to nonlinear models by Schraudolph \citep{schraudolph1999} using fast Hessian-vector product techniques. Using SMD to optimize neural network models is an ongoing area of research \citep{vivek2017,wu2018understanding,jacobsen2019meta,kearney2019learning}. It is an interesting question whether gradient sparsity emerges when applying SMD to online learning problems that are not clearly structured in tasks, as considered here. Furthermore, this line of work suggests that it might be possible to obtain finer binary gradient mask updates in an online fashion using forward-mode automatic differentiation.

A recent study has put into question whether any useful adaptation still takes place when MAML few-shot learners are presented with a novel task after meta-learning \citep[][]{raghu_rapid_2020}. Our findings shed light on this question, by demonstrating that few-shot learning performance can be improved when learning an adequate small subset of parameters. The additional plasticity of our meta-learned sparse learners led to a significant performance increase over handwired schemes based on frozen layers, in particular when encountering tasks drawn from a different family of problems than that used for meta-learning.

Our results may be of special interest to the design of neuromorphic hardware. Updating weights on-chip implies a significant power overhead whose cost scales with the number of plastic weights \citep[][]{park201965}. Reducing the number of plastic weights can therefore result in immediate improvements in energy efficiency and scalability. Likewise, synaptic plasticity is costly in biological neural networks. Given the high energy demands of the brain there has likely been selective pressure to reduce costs associated with synaptic change \citep[][]{li2020energy}. It is therefore conceivable that the brain developed mechanisms to restrict learning to an appropriate subset of synapses to save energy.  Our study presents further evidence in favor of sparse synaptic change, given its potential benefits in the biologically-relevant scenarios of few-shot and continual learning investigated here.

\paragraph{Limitations.} To arrive at our simple mask update (\eqref{eq:sp-MAML-mask-update}) we introduced two important approximations. First, we dropped all terms involving second-order derivatives, and second, we used straight-through estimation to differentiate through the step function. Despite their frequent use in previous work, both approximations remain poorly understood. For this reason it is possible that our algorithm fails unexpectedly outside the experiments considered here; potential problems stemming from the non-differentiability of the step function are likely unavoidable in our approach. Our update is also likely inappropriate for long inner loops (large $K$) \citep[][]{wu2018understanding}. Finally, scaling our few-shot learning experiments to more complex neural network models is potentially difficult, a challenge that our approach shares with other methods based on MAML. 

\begin{ack}
This work was supported by an Ambizione grant (PZ00P3\_186027) awarded to João Sacramento by the Swiss National Science  Foundation. Johannes von Oswald  is funded by the Swiss Data Science Center (J.v.O.~P18-03). Dominic Zhao is supported by AlayaLabs (Montreal, Canada). 
Massimo Caccia was supported through MITACS during his part time employment with Element AI the ServiceNow company, and by Amazon, during his part time employment there.
We thank Charlotte Frenkel, Frederik Benzing, Angelika Steger and Laura Sainz for helpful discussions.
\end{ack}

\bibliography{lw2l}
\bibliographystyle{plainnat}

\newpage
\normalsize
\setcounter{page}{1}
\setcounter{figure}{0} \renewcommand{\thefigure}{S\arabic{figure}}
\setcounter{table}{0} \renewcommand{\thetable}{S\arabic{table}}
\appendix

\section*{\Large{Supplementary Material\\\vspace{0.3cm}Learning where to learn: Gradient sparsity in meta and continual learning}}
\textbf{Johannes von Oswald*, Dominic Zhao*, Seijin Kobayashi, Simon Schug, Massimo Caccia, }\\
\textbf{Nicolas Zucchet, João Sacramento}\\\\

\section{Derivation of the sparse-MAML update}
\label{apx:sparse_maml}

Here, we derive the sparse-MAML update rules on the initialization $\theta$ and on the underlying mask parameter $m$, that are given by
\begin{align}
    \label{eqn:sparse-maml-update-theta}
    \theta & \gets \theta - \gamma_\theta \, \mathbb{E}_{\tau\sim p(\tau)}\!\left[\nabla_\phi \, \Lo_\tau\!\left(\phi_{\tau,K}\right)\right]\\
    \label{eqn:sparse-maml-update-mask}
    m & \gets m + \alpha \, \gamma_m \, \mathbb{E}_{\tau \sim p(\tau)} \!\left [ \nabla_\phi \, L_\tau^{\text{out}}\!\left(\phi_{\tau, K} \right) \circ \sum_{k=0}^{K-1} \nabla_\phi \, L_\tau^{\text{in}}\!\left( \phi_{\tau, k} \right) \right ]\!.
\end{align}

\paragraph{Update of the initialization}
We first start by deriving the $\theta$-update. To update $\theta$ with gradient descent we need the total derivative $\text{d}_\theta \, L^\text{out}_\tau(\phi_{\tau, K})$. Using the chain rule, it is equal to
\begin{align*}
    \text{d}_\theta \, L^\text{out}_\tau(\phi_{\tau, K}) &= \nabla_\phi \, L^\text{out}_\tau(\phi_{\tau, K}) ~ \text{d}_\theta \, \phi_{\tau, K}.
\end{align*}
The last term of the right hand side of the previous equation requires backpropagating through the training procedure as modifying the initialization changes the entire trajectory of $\phi$. By using the recursive formulation of $\phi_{\tau, K}$, we have
\begin{align*}
    \text{d}_\theta \, \phi_{\tau, K} &= \text{d}_\theta \! \left [ \phi_{\tau, K-1} - \alpha \mathbbm{1}_{m\geq 0} \circ \nabla_\phi \, L^\text{in}_\tau(\phi_{\tau, K-1}) \right ]\\
    &= \text{d}_\theta \, \phi_{\tau, K-1} - \alpha \mathbbm{1}_{m\geq 0} \circ \left ( \nabla_\phi^2 \, L^\text{in}_\tau(\phi_{\tau, K-1}) ~ \text{d}_\theta \, \phi_{\tau, K-1} \right)\!.
\end{align*}
In sparse-MAML, we use a first-order approximation that consists in zeroing out all the second order derivatives to keep the computations as simple as possible, while keeping the benefits of meta-learning. It follows that
\begin{align*}
    \text{d}_\theta \, \phi_{\tau, K} &\approx \text{d}_\theta \, \phi_{\tau, 0}\\
    &= \text{d}_\theta \, \theta\\
    &= \text{Id}
\end{align*}
and 
\begin{equation*}
    \text{d}_\theta \, L^\text{out}_\tau(\phi_{\tau, K}) \approx \nabla_\phi \, L^\text{out}_\tau(\phi_{\tau, K}),
\end{equation*}
leading to the update presented in Eq.~\ref{eqn:sparse-maml-update-theta} once the derivative approximation is inserted in a gradient descent update.

In our online continual learning setting, we additionally apply the mask to the $\theta$-update.

\paragraph{Update of the mask}
The derivation of the underlying mask parameter $m$ update can be done similarly to the one of the $\theta$-update. We first apply the chain rule and get
\begin{equation*}
    \mathrm{d}_m \, L_\tau^\text{out}(\phi_{\tau,K}) = \nabla_\phi \, L^\text{out}_\tau(\phi_{\tau,K}) ~ \mathrm{d}_m \phi_{\tau,K}.
\end{equation*}
We then compute the derivative of $\phi_{\tau,K}$ with respect to $m$:
\begin{equation*}
    \mathrm{d}_m \, \phi_{\tau, K} =  \mathrm{d}_m \, \phi_{\tau, K-1} - \alpha \, \mathrm{d}_m \! \left [ \mathbbm{1}_{m\geq 0} \circ \nabla_\phi \, L^\text{in}_\tau( \phi_{\tau, K-1}) \right ]\!.
\end{equation*}
As for the $\theta$-update, we do not take in account second-order derivatives, we thus consider first-order derivatives to be constant. The following terms remain
\begin{equation*}
    \mathrm{d}_m \, \phi_{\tau, K} \approx \mathrm{d}_m \, \phi_{\tau, K-1} -\alpha \, \mathrm{d}_m\left [ \mathbbm{1}_{m\geq 0} \right ] ~ \mathrm{diag}\!\left(\nabla_\phi\, L^\text{in}_\tau (\phi_{\tau, K-1})\!\right)\!.
\end{equation*}
We approximate $\mathrm{d}_m \mathbbm{1}_{m\geq 0}$ using straight-through estimation, which consists in taking this derivative equal to the identity, thus having
\begin{equation*}
    \mathrm{d}_m \, \phi_{\tau, K} \approx \mathrm{d}_m \phi_{K-1} -\alpha \, \mathrm{diag}\!\left(\nabla_\phi L^\text{in}_\tau ( \phi_{\tau, K-1})\!\right)
\end{equation*}
and
\begin{equation*}
    \mathrm{d}_m \, \phi_{\tau, K} \approx -\alpha \sum_{k=0}^{K-1} \mathrm{diag}\!\left(\nabla_\phi L_\tau^\text{in}( \phi_{\tau, k})\!\right)\!.
\end{equation*}
Combining everything into a gradient descent update yields the update of Eq.~\ref{eqn:sparse-maml-update-mask}.

Note that the updates for $\theta$ and $m$ differ in their structure although both are obtained using first-order approximations. This is because $\theta$ only enters the first update step of $\phi$, while $m$ consistently appears along the whole trajectory of $\phi$.

\section{Additional experimental details and analyses}

\subsection{Few-shot learning experiments}
\subsubsection{Reproducibility}

Unless specified otherwise, all experiments presented in our paper follow the supervised few-shot learning setup studied in  ref.~\citep{finn_model-agnostic_2017} and are performed on the miniImageNet dataset \citep{ravi_2016,vinyals_matching_2017} which consists of 64 training classes, 12 validation classes and 24 test classes. 
The backbone classifier consists of four convolutional layers each with 64 filters followed by a batch normalization layer \citep{ioffe_batch_2015} as well as a max-pooling layer with kernel size and stride of 2. The network then projects to its output via a fully-connected layer. We choose to use the 64-filter version (instead of the 32) to be one-to-one comparable to BOIL \cite{oh2021boil} (and the ANIL results within) which uses the 64 channel variant. 

In order to produce the results visualized in Figures~\ref{fig:inner_layer_sparse} and \ref{fig:number_of_steps}, we used the following hyperparameters: 
\begin{itemize}
    \item Batch size 4 and 2 for 1-shot resp.~5-shot experiments (note that BOIL uses 4 for both).
    \item Inner-loop length $K = 25$ during meta-training and meta-test train.
    \item Inner-loop learning rate $\alpha=0.1$.
    \item Optimizer: Adam with default PyTorch hyperparameters and a learning rate of 0.001 (for meta-parameters $\theta$ and $m$).
    \item Initialization: Kaiming \citep{he_delving_2015} for meta-parameters $\theta$ and $m$. 
\end{itemize}
Note that when analyzing the effects of varying a particular set of hyperparameters (e.g., the inner-loop learning rate), we hold all other hyperparameters fixed.

We train all models for 400 epochs (600 for sparse MAML$^+$) of 100 training tasks each. In the case of sparse-ReLU we use the initialization proposed in Meta-SGD \cite{DBLP:journals/corr/LiZCL17} and uniformly sample weights from the interval of $[0.05, 0.1]$. Note that this leads to an initial gradient sparsity level of $0\%$, while still converging to high sparsity levels.
%We refer to Table~\ref{Tab:sparsity_table} for an overview of network-averaged gradient sparsity levels of the final models, for all sparse-MAML variants.

All our few-shot learning results are reported for models that are early-stopped by measuring the average validation set accuracy (across 300 validation set tasks). The model with best average validation set accuracy is then tested on 300 tasks of the test set data and the cross-domain datasets. 

We handle batch normalization parameters following the \textit{transductive} learning setting, as originally done in MAML \citep{finn_model-agnostic_2017, nichol_first-order_2018}. 
\label{apx:exper_section}

For the results shown in Table \ref{Tab:results_table}, we tuned the best values found by scanning over learning rates and inner-loop lengths using a sparsity initialization of $50\%$. Additional details can be found in Table~\ref{Tab:few_shot_hyp}.

\paragraph{ResNet-12}
For the ResNet-12 results shown in Table \ref{Tab:resnet_results_table}, we tuned the best values found by scanning over inner-loop learning rates and inner-loop lengths with a sparsity initialization of $50\%$. For sparse-ReLU-MAML we initialized the inner-loop learning rate to be $\alpha$ without any randomness. For all experiments we optimize meta-parameters with Adam \cite{kingma_adam_2015}. We also set $\gamma_\theta =0.001, \alpha=0.05$ and $\gamma_m=0.01, K_{\text{test/train}} = 35$. The architecture is identical to the one used in previous meta-learning studies \cite{oh2021boil,meta_opt_net}.
We tested two different sizes for the ResNet that we term \emph{large}, with channel sizes $(64, 160, 320, 640)$, and \emph{small}, with channel sizes $(64, 128, 256, 512)$. We also adapted a strategy by \cite{DBLP:conf/iclr/AntoniouES19} where we test on an ensemble of the $3$ best models checkpointed while training. The results of all these variants are shown in Table~\ref{Tab:all_resnet_12_results}.

\begin{table}
\centering
  \caption{Detailed results for the large and small ResNet-12 models on miniImageNet 5-way 1- and 5-shot experiments, including sparsity levels and the performance of an economical snapshot ensemble method used in previous studies of MAML \citep{DBLP:conf/iclr/AntoniouES19}. Mean $\pm$ std.~over 3 seeds.} 
  \label{Tab:all_resnet_12_results}
   \begin{tabular}{lllccc}
    \toprule
  Arch. & Problem & Method  &  Acc. ($\%$)  & Ensemble Acc. ($\%$) & Sparsity ($\%$) \\
    \midrule
    \multirow{8}{*}{Large}
   &  \multirow{4}{*}{1-shot}
    & MAML  &  53.51$^{\pm1.24}$  & 55.65$^{\pm0.81}$  & ---  \\
   & & ANIL  &  52.95$^{\pm1.30}$  & 55.23$^{\pm0.66}$ & \text{all except head} \\
   & & sp-M   &  55.18$^{\pm0.50}$  &  56.83$^{\pm0.08}$  &  48.39 \\
   & & sp-ReLU-M & 55.29$^{\pm0.56}$  &  57.44$^{\pm0.43}$  &  29.56 \\
   \cmidrule(lr{1em}){2-6}
   & \multirow{4}{*}{5-shot} 
     & MAML  &  69.58$^{\pm1.08}$  & 72.77$^{\pm0.60}$  & ---  \\
   & & ANIL  &  69.39$^{\pm1.28}$  & 73.07$^{\pm0.42}$ & \text{all except head} \\
   & & sp-M   & 69.93$^{\pm0.61}$  & 72.83$^{\pm0.35}$ 	  &  23.57 \\
   & & sp-ReLU-M & 72.93$^{\pm0.92}$  &  75.60$^{\pm0.12}$  &  12.95 \\
       \midrule
    \multirow{8}{*}{Small}
   &  \multirow{4}{*}{1-shot}
    & MAML  &  53.91$^{\pm0.61}$  & 56.09$^{\pm0.12}$  & ---  \\
   & & ANIL  &  55.25$^{\pm0.33}$  & 57.02$^{\pm0.21}$ & \text{all except head} \\
   & & sp-M   &  55.02$^{\pm0.46}$  &  57.53$^{\pm0.25}$  &  37.56 \\
   & & sp-ReLU-M & 56.39$^{\pm0.38}$  &  58.41$^{\pm0.38}$  &  28.44 \\
   \cmidrule(lr{1em}){2-6}
   & \multirow{4}{*}{5-shot} 
     & MAML  &  69.36$^{\pm0.23}$  & 72.50$^{\pm0.22}$  & ---  \\
   & & ANIL  &  70.03$^{\pm0.58}$  & 73.09$^{\pm0.13}$ & \text{all except head} \\
   & & sp-M   & 70.02$^{\pm1.12}$  & 72.87$^{\pm0.59}$  & 15.09 \\
   & & sp-ReLU-M & 73.01$^{\pm0.24}$  &  75.52$^{\pm0.48}$  &  15.78 \\
    \bottomrule
  \end{tabular}
\end{table}

\begin{table}
\centering
  \caption{Hyperparameters of sparse-MAML,  sparse-MAML$^+$ and sparse-ReLU-MAML to obtain the reported results for in- and cross dataset few-shot experiments.} 
  \label{Tab:few_shot_hyp}
   \begin{tabular}{lllllllc}
    \toprule
    Problem & Method  & Optimizer & $K_{\text{train}}$  &  $K_{\text{test}}$ & $\alpha$ & $\gamma_m$ & $K_{\text{test}}^{\text{tiered}}$/$K_{\text{test}}^{\text{CUB}}$/$K_{\text{test}}^{\text{Cars}}$\\
    \midrule
    
    \multirow{3}{*}{1-shot}
    & sp-M   &  Adam & 35 &100 & 0.1 & 0.0075& 35\\
    & sp-M$^+$ & SGD+N  & 35 & 100 & 0.1 & 0.0075& 35\\ 
    & sp-ReLU-M  & Adam & 35 & 100& 0.1 & 0.001& 35\\ 

    \midrule
    \multirow{3}{*}{5-shot} 
    & sp-M  & Adam & 35 & 100& 0.25 & 0.0075& 100\\ 
    & sp-M$^+$ & SGD+N  & 35 & 100& 0.1 & 0.0075& 100\\ 
    & sp-ReLU-M & Adam & 35 & 100& 0.5 & 0.005& 100\\ 
    \bottomrule 
  \end{tabular}
\end{table}

\paragraph{Sparse-MAML$^+$.} To generate the underlying mask parameter $m\in \mathbb{R}^{N}$ in sparse-MAML$^+$ ($N$ being the dimension of the parameter space) we apply an affine transformation to a Gaussian vector $z \in \mathbb{R}^E$ (we set $E=1600$) with explicitly learnable noise standard deviation $\sigma$:
\begin{equation}
    m = A \, (z \circ \sigma + \mu)+ b,
\end{equation}
with $A \in \mathbb{R}^{N\times E}$, $b \in \mathbb{R}^{N}$, $\sigma,\mu \in \mathbb{R}^E$ and $z \sim \mathcal{N}(0, I)$. We use this process to generate the gradient mask parameters for convolutional layers only. As with every variant of sparse-MAML studied here, we adjust the meta-parameters $A, b, \mu, \sigma$ using a first-order update and straight-through estimation.

\subsubsection{Additional analyses}
\begin{figure}[h!]
\centering
\hspace{-10pt}
\begin{minipage}{.50\textwidth}
  \centering
  \begin{center}
    \includegraphics[width=0.8\textwidth]{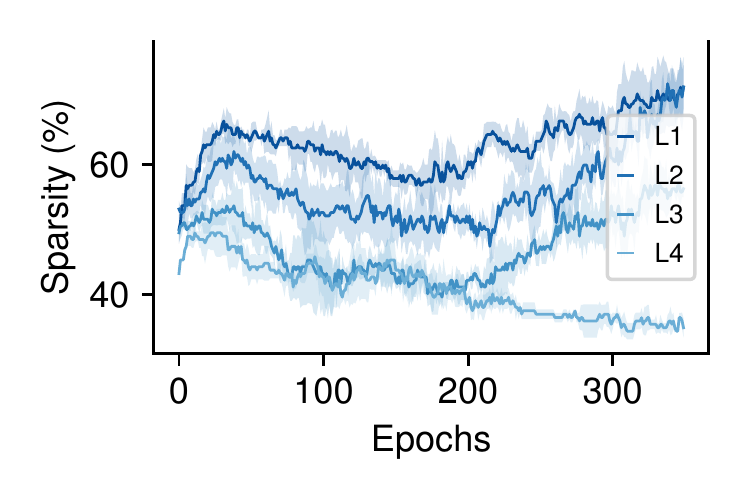}
  \end{center}
  \vspace{-15pt}
\end{minipage}
\hspace{-12pt}
\begin{minipage}{.50\textwidth}
  \centering
  \begin{center}
    \includegraphics[width=0.8\textwidth]{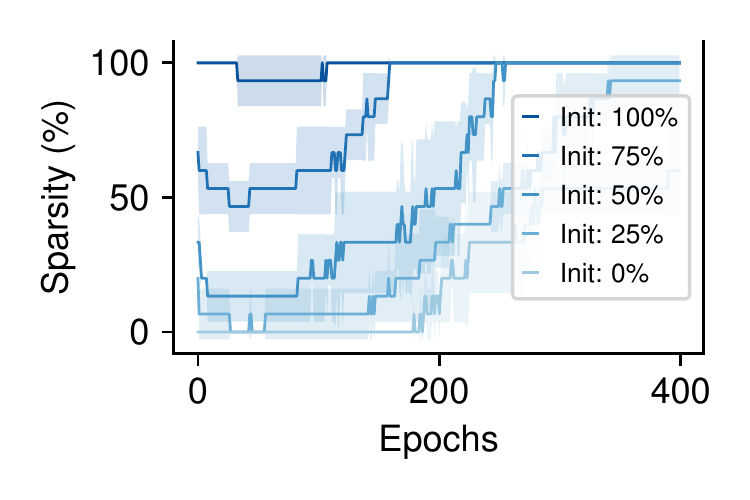}
  \end{center}
  \vspace{-15pt}
\end{minipage}
\hspace{-12pt}
  \captionof{figure}{Emergent gradient sparsity in 5-shot 5-way classification of miniImageNet on the standard 4-convolutional-layer neural network, with inner-loop learning rate $0.1$ and $25$ inner-loop steps. Results averaged over 5 seeds $\pm$ std. \textit{Left}: Different final gradient sparsity for batch normalization gain parameters emerges with gradually less sparsity from earlier to deeper layers, all initialized at $~50\%$ sparsity. \textit{Right:} Output layer bias parameter sparsity for different initial sparsity levels tend towards $100\%$. Note that deeper layers typically tend towards lower levels of sparsity.}
\label{apx:inner_layer_sparse2}
\vspace{-5pt}
\end{figure}

Complementing Figure \ref{fig:inner_layer_sparse}, we show in Figure \ref{apx:inner_layer_sparse2} emerging gradient sparsity in batch normalization and bias parameters throughout the network. Interestingly, we observe non-monotonicity in the sparsity levels especially in batch normalization parameters throughout training. This is possible by allowing to change sparsity in both directions by using the straight-through estimator for the binary mask. We find that the bias parameters eventually become entirely frozen (Figure~\ref{apx:inner_layer_sparse2} right) irrespective of initialization.

\begin{figure}[h!]
\centering
\hspace{-10pt}
\begin{minipage}{.5\textwidth}
  \centering
  \begin{center}
    \includegraphics[width=0.9\textwidth]{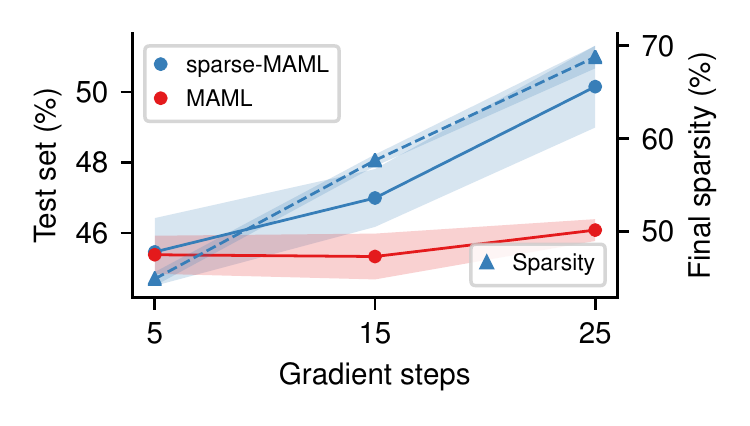}
  \end{center}
  \vspace{-15pt}
\end{minipage}
\hspace{-12pt}
\begin{minipage}{.5\textwidth}
  \centering
  \begin{center}
    \includegraphics[width=0.9\textwidth]{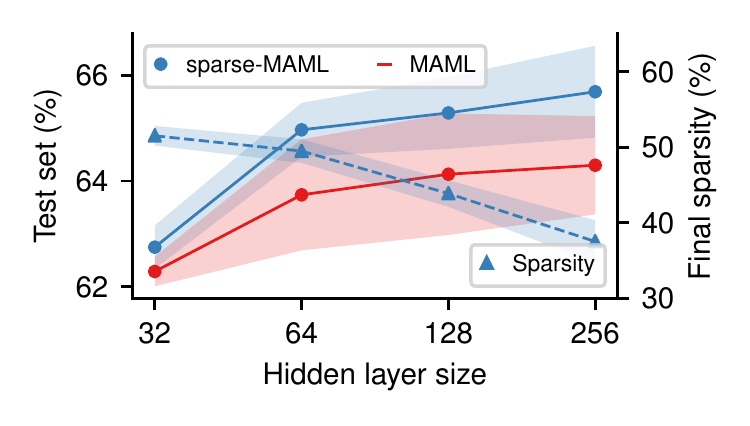}
  \end{center}
  \vspace{-15pt}
\end{minipage}
  \captionof{figure}{Additional results for 5-way miniImageNet classification, complementing Figure~\ref{fig:number_of_steps}. \textit{Left}: In 1-shot learning problems, long inner-loops lead to an increase in generalization performance accompanied by high gradient sparsity levels. By contrast, the performance of standard MAML does not improve with longer inner-loops. \textit{Right:} Gradient sparsity decreases as hidden layer width increases. The inner-loop learning rate was set to $0.1$ for all hidden layer sizes and gradient sparsity is initially $\sim50\%$. Results are averages over 5 seeds $\pm$ std.}
  \label{fig:1shot-and-width}
\vspace{-8pt}
\end{figure}
We additionally carry out an analysis of models with varying hidden layer sizes, cf.~Figure~\ref{fig:1shot-and-width}, and find that sparsity is anti-correlated with network width, indicating that the pressure of preventing interference by sparse gradients is reduced in large-capacity models.

\begin{table}
\centering
  \caption{Additional 5-way 5-shot miniImageNet few-shot learning experiments investigating the non-transductive batch normalization setting, and an ablation study in which bias parameters (which are consistently frozen by sparse-MAML) are removed from the model.} 
  \label{Tab:bias_transductive}
   \begin{tabular}{lc}
    \toprule
    Algorithm & Test set acc. ($\%$) \\
    \midrule
    sparse-MAML & 67.03$^{\pm0.74}$  \\
    sparse-MAML w/o bias parameters & 66.11$^{\pm0.57}$  \\
    FOMAML non-transductive BatchNorm & 55.58$^{\pm1.68}$  \\
    sparse-MAML non-transductive BatchNorm & 54.75$^{\pm1.17}$  \\
    \bottomrule
  \end{tabular}
\end{table}
We further show the performance of sparse-MAML on models without bias parameters, as these are consistently chosen to be frozen by meta-learning, to verify whether they are useful as task-shared parameters or simply not required at all. We find that performance drops slightly when removing the bias parameters, Table~\ref{Tab:bias_transductive}, which indicates that sparse-MAML ascribes to these parameters the role of providing useful task-shared bias.

These experiments are complemented by a study of the challenging non-transductive BatchNorm setting. Here, we simply compute batch statistics over the course of meta-train/test training without computing new statistics during meta-train/test testing -- we point to ref.~\cite{tasknorm} for a discussion. Since FOMAML was close to chance-level performance for $35$ inner-loop steps, the results reported for FOMAML are produced with $10$ inner-loop steps and $\alpha=0.1$. Sparsity emerges again with sparse-MAML, although now without bringing a performance advantage over FOMAML, see Table~\ref{Tab:bias_transductive}. All hyperparameters were kept the same as described in Table \ref{Tab:few_shot_hyp}, except for the change in inner-loop length and $\alpha$ that was needed to stabilize FOMAML.

We present one last few-shot learning study in Table~\ref{Tab:post_training_sparsity}, where we test whether meta-learning of the model initialization $\theta$ and the sparsity mask $m$ have to happen jointly, or if an appropriate gradient mask can be found separately after standard MAML training, keeping $\theta$ fixed. We find that this form of meta-learning fails to improve upon standard MAML alone. Thus, the generalization performance improvements brought by sparse-MAML rely on discovering a model initialization that is specialized for sparse learning. These results indicate that it is unlikely that the performance of sparse-MAML can be reached by simply analyzing the MAML solution post-training and heuristically disabling certain weight updates.

\begin{table}
\centering
  \caption{Two-phase learning experiments: meta-learning a gradient mask after learning the model initialization using standard MAML does not result in improved generalization performance on 5-shot miniImageNet learning.} 
  \label{Tab:post_training_sparsity}
   \begin{tabular}{lccc}
    \toprule
    Setup & Initial sparsity ($\%$) & Final sparsity ($\%$) & Test set acc. ($\%$) \\
    \midrule
    1-shot sparse-MAML & 0 & 9 & 46.42$^{\pm0.58}$  \\
    1-shot sparse-MAML & 50 & 45& 46.42$^{\pm0.27}$  \\
    5-shot sparse-MAML & 0 &29 &64.68$^{\pm0.16}$  \\
    5-shot sparse-MAML & 50 & 51& 64.01$^{\pm0.47}$\\
    \bottomrule
  \end{tabular}
\end{table}

\begin{wrapfigure}[15]{R}{0.52\textwidth}
\vspace{-0.8cm}
    \begin{minipage}{0.52\textwidth}
    \begin{algorithm}[H]
      \KwRequire{Parameters $\theta$, mask parameters $m$, replay buffer $R$, incoming batch of data $\mathcal{B}$, inner-loop learning rate $\alpha_0$, mask learning rate $\gamma_m$, loss $\mathcal{L}$}

      $\phi \gets \theta$
      
      $g^\text{in} \gets 0$
      
      \For{$1 \le k \le |\mathcal{B}|$}{
    
       $\phi \gets \phi - \alpha_0 \, \mathbbm{1}_{m\geq 0} \circ \nabla \mathcal{L}(\phi, \mathcal{B}_k)$
       
       $g^\text{in} \gets g^\text{in} + \nabla \mathcal{L}(\phi, \mathcal{B}_k)$
       
      }
      $\mathcal{R} \gets$ Sample past data batch from $R$
      
      $m \gets m + \gamma_m \, \alpha_0 \, \nabla \mathcal{L}(\phi, \mathcal{B} \cup \mathcal{R}) \circ g^\text{in}$
      
      $\theta \gets \theta - \alpha_0 \, \mathbbm{1}_{m\geq 0} \circ \nabla \mathcal{L}(\phi, \mathcal{B} \cup \mathcal{R})$
      
      $R \gets$ Update replay buffer $R$ with $\mathcal{B}$
      \caption{One step of sparse-La-MAML\label{alg:sp-lam}}
     \end{algorithm}
    \end{minipage}
\end{wrapfigure}

\subsection{La-MAML experiments}
\label{apx:la_maml}
\subsubsection{Reproducibility}

We strictly follow the experimental setup of the original La-MAML study and use the code provided by the authors\footnote{\url{https://github.com/montrealrobotics/La-MAML}}. The reported results are obtained by scanning three hyperparameters in the same range considered in the original paper, cf.~Appendix of ref.~\cite{gupta_-maml_2020}. Therefore, we only vary the number of glances within $\{5, 10\}$,  the outer-loop mask learning rate $\gamma_m$, and the inner-loop learning rate $\alpha_0$. See Table \ref{Tab:la_maml_hyp} for the hyperparameters found by our scan. 
The network used for the MNIST experiments is a 2-hidden-layer neural network with 100 hidden rectified linear units and the number of parameters is 89.610: [(78400, 100), (10000, 100), (1000, 10)] in (no.~of weights, no.~of biases) format and in input-to-output order. The output layer has a softmax nonlinearity and we use the cross-entropy loss.

Pseudocode for one complete iteration of sparse-La-MAML can be found in Algorithm~\ref{alg:sp-lam}. The fixed-size replay buffer $R$ is updated stochastically with the reservoir sampling method presented in ref.~\citep[][]{riemer2019learning}.
\begin{wrapfigure}[16]{R}{0.5\textwidth}
\vspace{-0.5cm}
  \begin{center}    
    \includegraphics[width=0.4\textwidth]{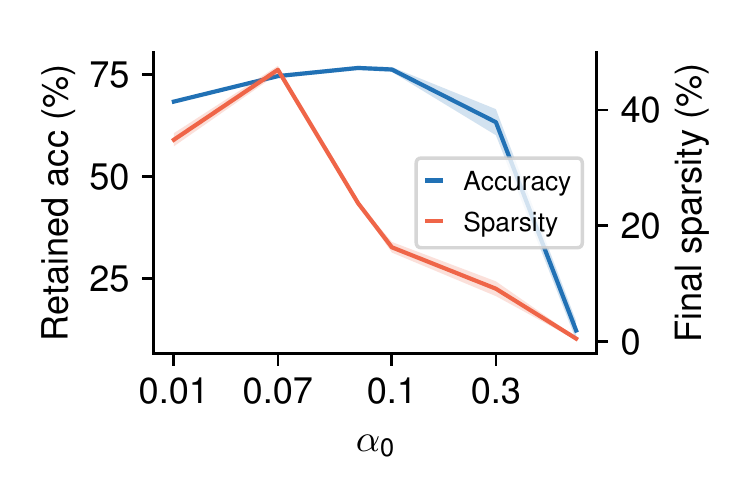}
  \end{center}
  \captionof{figure}{Retained accuracy (RA) and final gradient sparsity levels (in \%) for sparse-La-MAML applied to MNIST permutations, for different settings of the inner-loop learning rate $\alpha_0$.}
  \label{fig:alpha_la_maml}
\end{wrapfigure}

Following the original La-MAML experimental setup, we study three supervised continual learning (CL) problems based on MNIST. In MNIST \emph{rotations} (20 tasks, 1000 examples per task), each task is a classification problem where MNIST digits rotated by a fixed task-specific common angle (in [0, $\pi$]) are to be classified.  In MNIST \emph{permutations} (20 tasks, 1000 samples per task) and the harder \emph{many permutations} variant (100 tasks, 200 examples per task), a fixed task-specific pixel shuffling order is applied to every MNIST digit instead.

To produce the results in the left panel of Figure~\ref{fig:lamaml_sparsity}, we choose the configuration used for MNIST rotations found by our scan (cf.~\ref{Tab:la_maml_hyp}) and vary the layer size of the two hidden layers of the fully-connected network. For the results shown in Figure~\ref{fig:la-maml-multipass-sparsity}, we keep the hyperparameters of the original La-MAML paper but iterate over the dataset 10 times (epochs) instead of only once.

\subsubsection{Additional analyses}
\begin{figure}[h!]
\centering
\hspace{-10pt}
\begin{minipage}{.35\textwidth}
  \centering
  \begin{center}
    \includegraphics[width=0.9\textwidth]{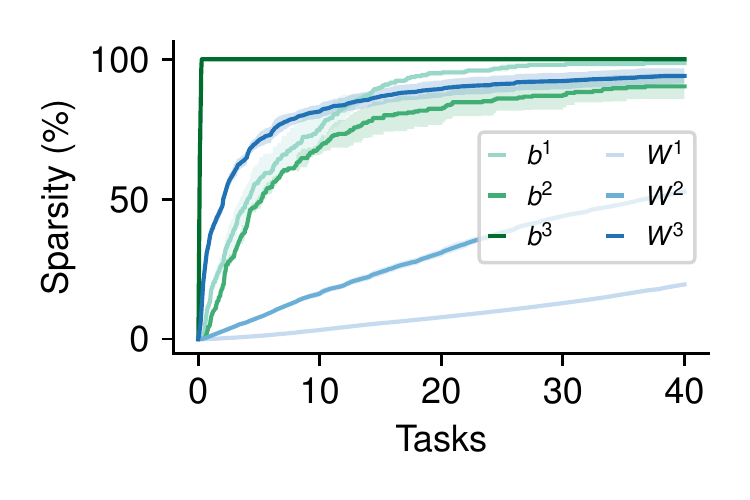}
  \end{center}
  %\vspace{-15pt}
\end{minipage}
\hspace{-12pt}
\begin{minipage}{.35\textwidth}
  \centering
  \begin{center}
    \includegraphics[width=0.9\textwidth]{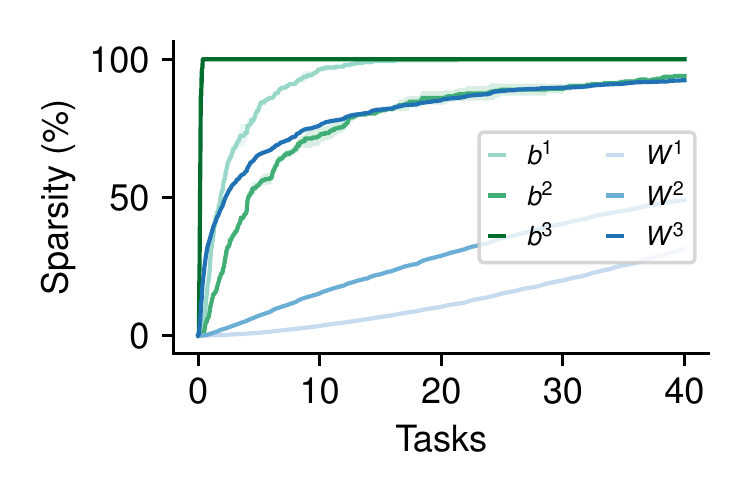}
  \end{center}
  %\vspace{-15pt}
\end{minipage}
\hspace{-14pt}
\begin{minipage}{.35\textwidth}
  \centering
  \begin{center}
    \includegraphics[width=0.9\textwidth]{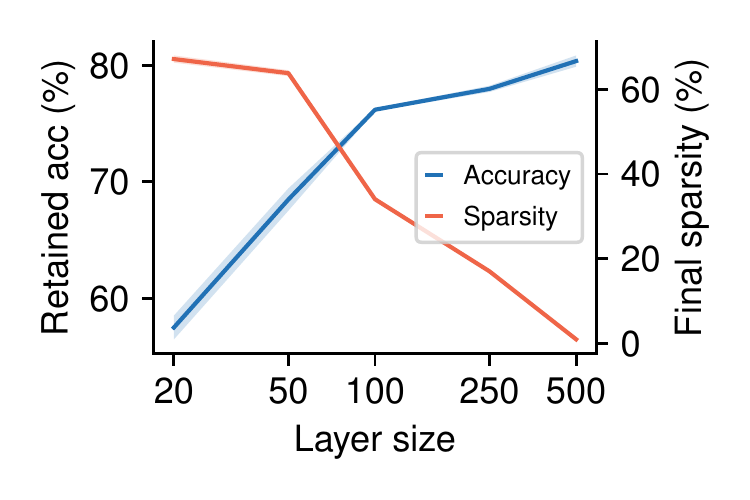}
  \end{center}
  %\vspace{-15pt}
\end{minipage}
\hspace{-12pt}
\begin{minipage}{.35\textwidth}
  \centering
  \begin{center}
    \includegraphics[width=0.9\textwidth]{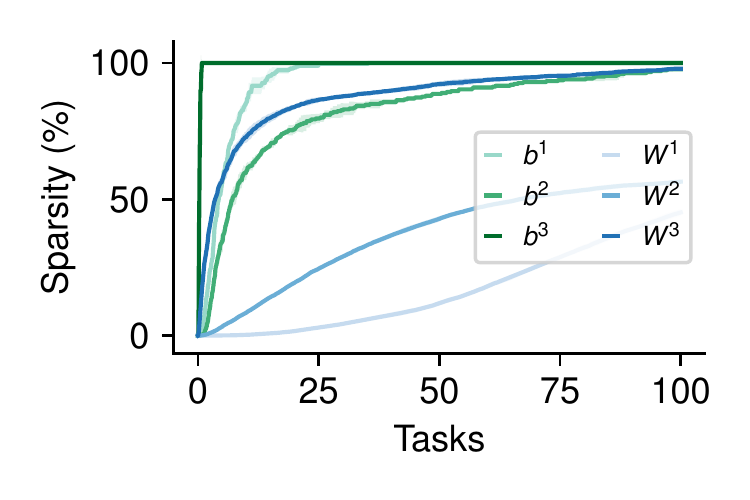}
  \end{center}
  %\vspace{-15pt}
\end{minipage}
\hspace{-16pt}
\begin{minipage}{.35\textwidth}
  \centering
  \begin{center}
    \includegraphics[width=0.9\textwidth]{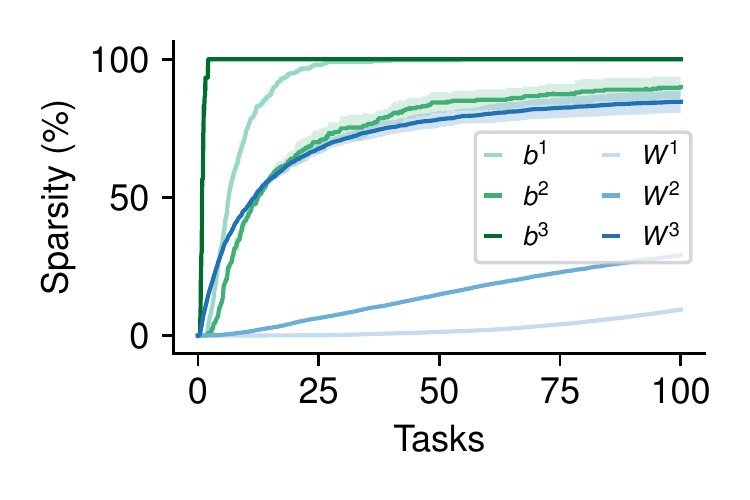}
  \end{center}
  %\vspace{-15pt}
\end{minipage}
\hspace{-16pt}
\begin{minipage}{.35\textwidth}
  \centering
  \begin{center}
    \includegraphics[width=0.9\textwidth]{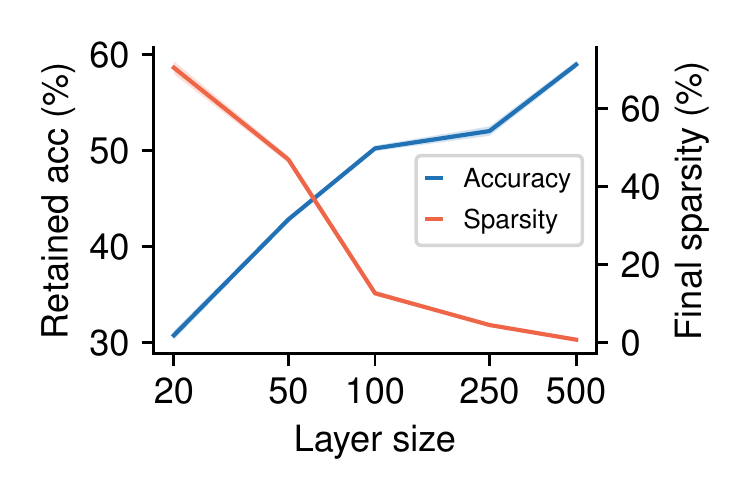}
  \end{center}
  %\vspace{-15pt}
\end{minipage}
  \captionof{figure}{Gradient sparsity when learning on MNIST permutations (upper column) and many permutations (lower column) with La-MAML (first row) and sparse-La-MAML (second and third row). Results averaged over 3 seeds $\pm$ std. \textit{Upper left (original La-MAML algorithm)}: Sparsity emerges across the three-layer network and monotonically increases with the number of tasks and with depth for both weight ($W_1, W_2, W_3$) and bias parameters ($b_1, b_2, b_3$). \textit{Center (sparse-La-MAML)}: A similar behavior is observed when replacing meta-learned learning rates by meta-learned binary gradient masks.  \textit{Right (sparsity/accuracy vs. layer size):} Overall sparsity of sparse-La-MAML  decreases with increased network capacity accompanied with higher retained accuracy (RA). Network capacity is varied by changing the number of neurons in the two hidden layers simultaneously.}
\label{fig:lamaml_sparsity_many}
%\vspace{-8pt}
\end{figure}

\begin{table}[b]
\vspace{-0.5cm}
\centering
  \caption{Sparsity ($\%$) of La-MAML and sparse-La-MAML after learning on one of the three MNIST continual learning problems rotations, permutations and many permutations. 
  Hyperparameters were tuned for best retained accuracy, not sparsity. We split between weight and bias parameters (weights followed by bias) when presenting per-layer average levels of sparsity within layers (ordered from network input to output). Structured gradient sparsity emerges, with lower-levels of sparsity for lower-level features closer to the input. Bias parameters tend to be close to frozen in almost all cases.} 
  \label{Tab:sparsity_values}
   \begin{tabular}{llll}
    \toprule
    Dataset &    Algorithm & Average within layers (\%) & Average (\%) \\
    \midrule
     \multirow{2}{*}{Rotations} & La-MAML & [16, 45, 96], [95, 94, 100] & 20.47 \\
     & sparse-La-MAML & [5, 19, 80], [67, 54, 100] & 7.13 \\ 
    %\midrule
    \cmidrule(lr{1em}){2-4}
    \multirow{2}{*}{Permutations} &  La-MAML & [20, 53, 97], [99, 93,  100] & 24.81 \\
     & sparse-La-MAML & [8, 29, 78], [100, 87, 100]& 16.17 \\ 
    %\midrule
    \cmidrule(lr{1em}){2-4}
    \multirow{2}{*}{Many permutations} &  La-MAML & [45, 56, 98], [100, 98, 100]& 46.53 \\
     & sparse-La-MAML & [10, 29, 87], [100, 93, 100] & 13.08 \\
    \bottomrule
  \end{tabular}
\end{table}
For completeness, we visualize the patterns of 
gradient sparsity that emerge when learning continually with La-MAML and sparse-La-MAML on the MNIST permutations and many permutations CL problems, see Figure~\ref{fig:lamaml_sparsity_many}. The findings reported in the main text translate to these two datasets, and the two variants of La-MAML again behave in a qualitatively similar way.

In our experiments, we observe that the inner-loop learning rate $\alpha_0$ has a strong effect on gradient sparsity. This is depicted in Figure~\ref{fig:alpha_la_maml} where the final gradient sparsity level for sparse-La-MAML trained on MNIST permutations is shown, together with retained accuracy. We find that while sparsity and accuracy are jointly maximized for lower inner-loop learning rate $\alpha_0$, high retained accuracies can still be achieved when increasing the learning rate $\alpha_0$, up to a point where accuracy eventually drops.

\begin{table}
\centering
  \caption{Hyperparameter settings for the reported La-MAML and sparse-La-MAML results.} 
  \label{Tab:la_maml_hyp}
   \begin{tabular}{llcccc}
    \toprule
    Dataset &  & Algorithm & $\alpha_0$ & $\gamma_m$ & $K$ / Glances \\
    \midrule
     \multirow{6}{*}{MNIST} &\multirow{2}{*}{Rotations}
     & LaM & 0.15 & 0.3 & 5 \\
     & & sp-LaM & 0.15 & 1.7 & 5 \\
    \cmidrule(lr{1em}){2-6}
    &\multirow{2}{*}{Permutations} & LaM & 0.15 & 0.3 & 5 \\
    & & sp-LaM & 0.1 & 1.7 & 10 \\
    \cmidrule(lr{1em}){2-6}
    &\multirow{2}{*}{Many} & LaM & 0.1 & 0.3 & 10 \\
    & & sp-LaM & 0.05 & 0.75 & 10 \\
    \bottomrule
  \end{tabular}
\end{table}

\begin{table}
\centering
  \caption{Final full CIFAR-10 test-set classification accuracy, continually-learned in a class-incremental, streaming fashion, in 5 tasks comprising 2 classes each. Each data point is seen only once. We compare sparse-La-MAML (sp-LaM; binary gradient masks, straight-through update), standard La-MAML (LaM; rectified learning rates, meta-learned without straight-through update), experience replay, gradient episodic memory (GEM) and meta-experience replay (MER), for two different replay buffer sizes. Results are averages over 5 seeds $\pm$ std.} 
  \label{Tab:la_maml_cifar}
   \begin{tabular}{llcccc}
    \toprule
    Total memory size &  Algorithm & Final acc. ($\%$) \\
    \midrule
     \multirow{5}{*}{200} & {Experience replay} & 19.75$^{\pm1.23}$  \\
     & MER & 25.11$^{\pm1.77}$  \\
      & GEM & 25.14$^{\pm0.67}$  \\
     & LaM & 22.08$^{\pm1.83}$ \\
     &  sp-LaM & 27.85$^{\pm0.69}$ \\
         \midrule

    \multirow{5}{*}{1000} & {Experience replay}& 29.12$^{\pm2.41}$  \\
     & MER & 34.66$^{\pm1.38}$  \\
      & GEM & 31.55$^{\pm0.81}$ \\
     & LaM &  36.24$^{\pm0.91}$ \\
     &  sp-LaM &  37.70$^{\pm0.80}$ \\
    \bottomrule
  \end{tabular}
\end{table}

\paragraph{Streaming Split-CIFAR-10 experiments.} Finally, we complement our continual learning investigation of gradient sparsity in La-MAML with results on a streaming Split-CIFAR-10 class-incremental learning problem. In this problem, the CIFAR-10 dataset is split into 5 tasks of 2 classes each, and each data point is processed online only once. We use a 4-convolutional-layer neural network, the same used in the original La-MAML paper \citep{gupta_-maml_2020}. This is a challenging setting where experience replay (ER) remains a strong baseline \citep{aljundi2019online}. We produced this baseline for our architecture, and compared sparse-La-MAML to it, performing for both methods a hyperparameter scan over replay batch size, the number of gradient updates per incoming batch, and learning rates. We also compare to GEM (while optimizing the following hyperparameters: batch sizes, number of gradient updates per batch, gradient clipping norm, the strength with which the memory constraint is enforced) and MER (scanning over batch sizes, regularization strength, gradient clipping norm, and learning rates).

For both small (20 examples per class) and large (100 examples per class) replay buffer sizes, sparse-La-MAML consistently outperforms ER, cf.~Table~\ref{Tab:la_maml_cifar}, as well as MER and GEM.

We observe a qualitatively distinct gradient sparsity pattern emerge in this setting, compared to our MNIST experiments. Here, sparse-La-MAML leads to a large fraction of frozen weights in the final fully-connected layer ($57.5 \pm 0.866 \%$ for the small replay buffer case, and $64.75 \pm 0.8292 \%$ for the large replay buffer), which increase as more classes are learned, and significantly lower values of gradient sparsity for the remaining parameters ($1.6 \pm 0.68 \%$ overall sparsity for the small replay buffer case, and $1.0 \pm 0.70 \%$ for the large replay buffer). Our results confirm that the strong performance of La-MAML translates to a more challenging class-incremental continual learning problem, and reveal that meta-learning finds a solution with large gradient sparsity in the final output layer.

We further compare to the original La-MAML implementation provided in \citep{lamamlcode} for the MNIST experiments, which uses rectified learning rates (Eq.~\ref{eq:la-maml-inner}) but not our straight-through update. As discussed in the main text, this leads to dead parameter updates that can never recover once the learning rate goes below zero. We find that this variant of La-MAML leads to very high levels of gradient sparsity in the entire model ($96.3 \pm 1.5 \%$ when using small replay buffers, and $34.0 \pm 1.39 \%$ when using large replay buffers) but a performance hit, highlighting the importance of fine-tuning gradient masks without aggressively shutting off learning.

We found the following hyperparameters to work the best for each particular method:
\begin{itemize}
    \item \emph{La-MAML}. Batch size and replay batch size: 10; number of gradient steps per data point: 2; For memory size 200, 1000 we used $\alpha_{\text{init}} = 0.005, 0.01$ and $\gamma = 0.1, 0.01$, resp.
    \item \emph{GEM}. Number of steps per data point: 2; memory strength: 0.5; 500 samples per task. For memory sizes of 200 and 1000, we use batch sizes of 20 and 5, resp.
    \item \emph{MER}. Batch size and replay batch size: 10, $\beta=0.1$. $\gamma=0.05$, $\gamma=0.08$ for memory sizes of 200 and 1000, resp.
    \item \emph{ER}. Batch size of 10 for both memory sizes. Gradient steps per data point: $2,4$, $\gamma$: 0.001, 0.01; replay batch size: $20, 10$, for memory sizes of 200 and 1000 resp.
\end{itemize}

\subsection{C-MAML experiments}

We provide the full performance overview of the different C-MAML variants studied in the main text together with related work in Table~\ref{tab:omniglot}.
\begin{wrapfigure}[13]{R}{0.4\textwidth}
\vspace{-0.3cm}
  \begin{center}
    \includegraphics[width=0.4\textwidth]{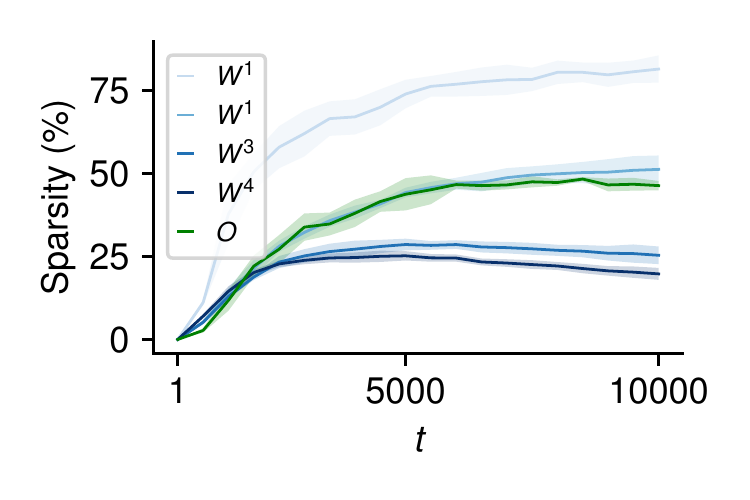}
  \end{center}
  \vspace{-0.5cm}
  %\vspace{-15pt}
  \captionof{figure}{Gradient sparsity when learning online with sparse-C-MAML. Gradient sparsity decreases with depth and rises again for the output layer.}
  \label{fig:sparsity-osaka}
\end{wrapfigure}

\setlength{\tabcolsep}{3.5pt}
\begin{table*}[b]
%   \centering
 \caption{Cumulative online accuracy on the Omniglot-MNIST-FashionMNIST online learning benchmark as well as the accuracy on the single tasks. Tasks switch with probability $1-p$. Results from previous work taken from ref.~\citep{caccia2020online}. Mean $\pm$ std.~over 5 seeds.}

  \centerline{
  \begin{scriptsize}
  \begin{sc}
    \begin{tabular}{l|cccc|cccc}
    & \multicolumn{4}{c|}{$p=0.98$} & \multicolumn{4}{c}{$p=0.90$} \\
    Method & Total & Omniglot & MNIST & Fashion & Total & Omniglot & MNIST & Fashion  \\ 
    \hline
        Online ADAM & 73.9  \std{2.2} & 81.7  \std{2.3} & 70.0  \std{3.6} & 62.3  \std{2.5} & 23.8  \std{1.2} & 26.6  \std{2.0} & 20.0  \std{1.4} & 22.1  \std{1.3} \\
        Fine Tuning & 72.7  \std{1.7} & 80.8  \std{2.0} & 68.7  \std{2.8} & 59.6  \std{3.1} & 22.1  \std{1.1} & 25.5  \std{1.5} & 18.1  \std{1.9} & 19.2  \std{1.6} \\
        MAML \citep{finn_model-agnostic_2017} & 84.5  \std{1.7} & 97.3  \std{0.3} & 80.4  \std{0.3} & 63.5  \std{0.3} & 75.5  \std{0.7} & 88.8  \std{0.4} & 68.1  \std{0.5} & 56.2  \std{0.4} \\
        ANIL \citep{raghu_rapid_2020} & 75.3  \std{2.0} & 95.1  \std{0.6} & 58.7  \std{2.9} & 49.7  \std{0.3} & 69.1  \std{0.8} & 88.3  \std{0.5} & 52.4  \std{0.6} & 47.6  \std{0.9} \\
        BGD \citep{DBLP:journals/corr/abs-2010-00373} & 87.8  \std{1.3} & 95.1  \std{0.5} & 86.9  \std{1.1} & 74.4  \std{1.1} & 63.4  \std{0.9} & 72.8  \std{1.2} & 55.9  \std{2.2} & 51.7  \std{1.3} \\
        MetaCOG  \citep{he2019task} & 88.0  \std{1.0} & 95.2  \std{0.5} & 87.1  \std{1.5} & 74.3  \std{1.5} & 63.6  \std{0.9} & 73.5  \std{1.3} & 55.9  \std{1.8} & 51.7  \std{1.4} \\
        MetaBGD  \citep{he2019task} & 91.1  \std{2.6} & 96.8  \std{1.5} & 92.5  \std{1.9} & 77.8  \std{3.8} & 74.8  \std{1.1} & 83.1  \std{1.0} & 71.7  \std{1.5} & 61.5  \std{1.2} \\
        \hline
        C-MAML   &  92.8  \std{0.6} &  97.8  \std{0.2} & 93.9  \std{0.8} & 79.9  \std{0.7} & 83.3  \std{0.4} & 89.0  \std{0.5} & 84.5  \std{0.7} & 71.1  \std{0.7} \\
        sparse-C-MAML  & 94.2 \std{0.4}  & 97.3 \std{0.1}  & 93.4\std{0.4} &  86.3\std{0.3} & 86.3 \std{0.4} & 89.3 \std{0.5} & 87.7\std{0.4} & 77.4\std{0.5} \\ 
        sparse-ReLU-C-MAML  & 93.5\std{0.5} & 97.16\std{0.2}  & 97.2\std{0.2} & 94.1\std{0.4} & 84.7 \std{1.3} & 89.3 \std{0.2} & 87.5 \std{0.5} & 78.3 \std{0.2}\\
    \end{tabular}
  \end{sc}
  \end{scriptsize}
  }
% \vspace{-3mm}
\label{tab:omniglot}
\end{table*}

\subsubsection{Reproducibility and ablation study}

In order to obtain the reported results, we use the code base\footnote{\url{https://github.com/ElementAI/osaka}} that accompanies ref.~\cite{caccia2020online}. We do not alter the architecture of the 4-convolutional-layer neural network (64 hidden units) used in the original C-MAML study. All our results are based on the best performing, non-ablated version of the C-MAML algorithm, termed C-MAML+UM+PAP in the original paper \citep[][]{caccia2020online}. Furthermore, we do not change the provided hyperparameters, and only tune the inner-loop and mask learning rates $\alpha_0$ and $\gamma_m$ (resp.) for our sparse-C-MAML and sparse-ReLU-C-MAML algorithm variants. For sparse-C-MAML in the $p=0.98$ setup, we initialized the mask parameters with the Kaiming initialization leading to an initial sparsity of $50\%$. For all sparse-ReLU-C-MAML runs, we initialized the mask parameters with a uniform initialization over the range $[0.005, 0.1]$.

We provide pseudocode for one iteration of sparse-C-MAML in Algorithm~\ref{alg:sp-c-maml}. Following Caccia et al.~\citep[][]{caccia2020online} and do not use any pretraining; the base parameters $\theta$ and the current parameters used for prediction $\phi$ are initialized randomly and equal to one another. The replay buffer $R$ is also initially empty. For details on the task change detection function and outer-loop learning rate adaptation function we refer to the original C-MAML study \citep[][]{caccia2020online}.

\paragraph{Omniglot-MNIST-FashionMNIST setup.} The Omniglot-MNIST-FashionMNIST benchmark studied here was introduced in the original C-MAML paper \citep[][]{caccia2020online}; we do not modify the experimental setup. In this online learning problem, at every time step $t$ the task changes with probability $1-p$. Each task is a $\mathcal{K}$-shot, 10-way classification problem. Tasks are created by sampling 10 classes uniformly (for Omniglot; MNIST and FashionMNIST are by default 10-way problems) and then sampling $\mathcal{K}$ examples for each of the selected 10 classes.

\paragraph{Gradient masking ablation study.} In order to verify the advantage of gradient masking, we also compare to an ablated version of C-MAML (called C-MAML-fixed) which does not feature any meta-learned learning rate parameters, setting the inner-loop learning rate to a fixed hyperparameter value (we note that the original C-MAML algorithm included a small set of meta-learned learning rates that were shared for large parameter groups and which were not restricted to be non-negative). Results are shown in Table \ref{Tab:osaka_hps}. In the Omniglot-MNIST-FashionMNIST experiment, the performance of C-MAML is matched by C-MAML-fixed.

In all of our experiments, we observed sparsity emerging and higher overall average accuracies for sparse-C-MAML compared to C-MAML-fixed and C-MAML. Note that the only difference between sparse-C-MAML and C-MAML-fixed is the ability to stop learning some of the parameters.

\begin{figure}
\centering
    \begin{minipage}{0.8\textwidth}
    \begin{algorithm}[H]
      \KwRequire{Current parameters $\phi$, meta-parameters $\theta$, mask parameters $m$, replay buffer $R$, incoming batch of data $\mathcal{B}$, inner-loop learning rate $\alpha_0$, mask learning rate $\gamma_m$, loss function $\mathcal{L}$, learning rate adaptation function $g$}

      \eIf{\textbf{\textup{not}} \text{task change detected}}{
      $\phi \gets \phi -  \alpha_0 \, \mathbbm{1}_{m\geq 0} \circ \nabla \mathcal{L}(\phi, \mathcal{B})$
      
      $R \gets R \cup \mathcal{B}$ \quad \tcp{\!update replay buffer with current data}
      }{
      $\mathcal{R}^\text{t} \gets$ sample batch of training data from $R$
      
      $\phi \gets \theta - \alpha_0 \, \mathbbm{1}_{m\geq 0} \circ \nabla \mathcal{L}(\theta, \mathcal{R}^\text{t})$
      
      $\mathcal{R}^\text{v} \gets$ sample batch of validation data from $R$
      
      $\eta \gets g(\mathcal{L}(\phi, \mathcal{R}^\text{v}))$ \quad \tcp{\!adapt learning rate}
      
      $\theta \gets \theta - \eta \, \nabla \mathcal{L}(\phi, \mathcal{R}^\text{v})$
      
      $R \gets \{\}$ \quad \tcp{\!reset replay buffer}
      
      $\phi \gets \theta -  \alpha_0 \, \mathbbm{1}_{m\geq 0} \circ \nabla \mathcal{L}(\theta, \mathcal{B})$
      }

      \caption{One step of sparse-C-MAML\label{alg:sp-c-maml}}
     \end{algorithm}
     \end{minipage}
\end{figure}

\section{Brief discussion on meta-learning-based approaches to continual learning}
The surge of meta-learning in continual learning can be explained by its ability to automatically discover the inductive biases that are appropriate for learning without forgetting. Previous works hypothesize that a particular inductive bias will mitigate catastrophic forgetting, e.g.~keeping parameters from diverging too much from previous versions \citep[][]{kirkpatrick2017}, and then develop a solution around that. Contrarily, meta-learning based approaches will learn inductive biases that are conducive for learning without interference in a data-driven way. For example, in ref.~\citep{javed2019meta} sparsity emerges in the learned representations, a characteristic that has long been hypothesized as desirable in continual learning \citep[][]{french_catastrophic_1999}.

Regularization-based methods are notoriously incapable of working in more realistic settings, such as those considered in our work, mostly because they are not equipped with a mechanism to perform cross-task discrimination or to recalibrate themselves on past tasks after some interference has occurred. The same applies for dynamic architectures, which rely on task labels. This reliance can be bypassed with a task-inference module, which may however suffer from some forgetting itself.

Rehearsal-based methods do not suffer from the aforementioned weaknesses. Nevertheless, they scale poorly due to their reliance on always approximating an i.i.d. distribution at every update. The total runtime of these methods scales quadratically with the number of tasks.

The ambitious goal of meta-learning inductive biases that benefit continual learning directly from data may come at the cost of decreasing sample efficiency and increasing compute requirements. However, the latter is potentially offset by reducing the number of hyperparameter-search trials \citep{maclaurin2015gradient}.

\section{Resources}

\paragraph{Compute.} We used 24 (3x8 servers) NVIDIA GeForce 2080 Ti GPUs for our experiments and conducted experiments and hyperparameter scans for approximately one month in order to obtain the reported results.

\begin{table}[h!]
\centering
  \caption{Task-averaged cumulative online accuracy of C-MAML, C-MAML-fixed and sparse-C-MAML and the hyperparameters that lead to the result.}
  \label{Tab:osaka_hps}
   \begin{tabular}{lllll}
    \toprule
    \cmidrule(r){1-2}
    & Method    & Accuracy (\%) & $\alpha_0$ & $\gamma_m$ \\
    \midrule
       \multirow{3}{*}{$p=0.9$} & C-MAML &83.3$^{\pm0.4}$ & 0.1 & 0.001 \\
    & C-MAML-fixed & 85.3$^{\pm0.5}$ & 0.3 & -\\
    & sparse-C-MAML &86.3$^{\pm0.4}$ & 0.3 & 0.003 \\ 
    & sparse-ReLU-C-MAML &86.1$^{\pm0.2}$ & - & 0.01 \\ 
    \midrule
    \multirow{3}{*}{$p=0.98$} & C-MAML &92.8$^{\pm0.6}$ & 0.1 & 0.005 \\
    & C-MAML-fixed & 92.0$^{\pm0.1}$ & 0.1 & - \\
    & sparse-C-MAML & 94.2$^{\pm0.4}$ & 0.3 & 0.01 \\ 
    & sparse-ReLU-C-MAML &93.5$^{\pm0.4}$ & - & 0.01 \\ 
    \bottomrule
  \end{tabular}
\end{table}

\paragraph{Software, libraries and licensing information.} The results reported in this paper were produced with open source, free software whenever possible. We developed custom code in Python using the PyTorch (BSD-style license) \cite{paszke_pytorch_2019} and NumPy (BSD-style license) \citep{harris2020array} libraries; few-shot learning dataset splits and meta-gradient computations further relied on the Torchmeta library (MIT license) version 1.6~\citep{deleu2019torchmeta}. Our extensions of the La-MAML (Apache-2.0 license) and C-MAML (unknown license; permission to extend granted by the authors) algorithms were built directly on top of the code distributed by the authors. All plots were generated using matplotlib (BSD-style license) \citep{hunter_matplotlib_2007}. Our computers run Ubuntu Linux.

We investigated our learning algorithms on the public domain datasets MNIST (GNU GPL v3.0) \citep[][]{lecun_mnist_1998}, FashionMNIST (MIT license) \citep[][]{xiao_fashion-mnist_2017}, Omniglot \citep[][]{lake_one_2011} (MIT license), miniImageNet \citep[][]{ravi_2016} (custom MIT/ImageNet license), CIFAR-10 (MIT license) \citep{krizhevsky_learning_2009}, CUB  (custom license) \citep[][]{welinder2010caltech}, tieredImageNet  (custom ImageNet license) \citep{ren2018meta} and Cars  (custom license) \citep[][]{KrauseStarkDengFei-Fei_3DRR2013}.

\section{PyTorch code snippet}
In all of our experiment we backpropagate through binary or ReLU masks using the straight-through estimator. For illustrative reasons, we provide a Python code snippet showing how to use either the ReLU straight-through or the binary mask e.g. inside an inner loop of MAML:

\newpage
%\DeclareFixedFont{\ttb}{T1}{txtt}{bx}{n}{10} % for bold
\DeclareFixedFont{\ttb}{T1}{txtt}{m}{n}{9} % for normal
\DeclareFixedFont{\ttm}{T1}{txtt}{m}{n}{9}  % for normal
\definecolor{deepblue}{rgb}{0,0,0.5}
\definecolor{deepred}{rgb}{0.6,0,0}
\definecolor{deepgreen}{rgb}{0,0.5,0}
\definecolor{deepyellow}{rgb}{1.,0.75,0}

% Default fixed font does not support bold face

\lstset{
language=Python,
basicstyle=\ttm,
morekeywords={self},              % Add keywords here
keywordstyle=\ttb\color{deepblue},
emph={Binary, ReLUThrough,__init__,@staticmethod},          % Custom highlighting
emphstyle=\ttb\color{deepred},    % Custom highlighting style
stringstyle=\color{deepgreen},
commentstyle=\color{deepgreen},
frame=tb,                         % Any extra options here
showstringspaces=false
}
\lstset{frame=lines}
\lstset{caption={Backpropagate through binary or ReLU mask}}
\lstset{label={lst:code_direct}}
\lstset{basicstyle=\footnotesize}
\begin{lstlisting}
import torch
class Binary(torch.autograd.Function):
    def __init__(self):
        super(Binary, self).__init__()
    @staticmethod
    def forward(ctx, input):
        return torch.sign(input)
    @staticmethod
    def backward(ctx, grad_output):
        return grad_output

class ReLUThrough(torch.autograd.Function):
    def __init__(self):
        super(ReLUThrough, self).__init__()
    @staticmethod
    def forward(ctx, input):
        return torch.relu(input)
    @staticmethod
    def backward(ctx, grad_output):
        return grad_output

def training():
    ...  
    #  Inside an inner loop
    grads = torch.autograd.grad(loss, weights)
    if ReLUThroughMask:
        params =  params - ReLUThrough.apply(m)*grads
    elif BinaryMask:
        # alpha is the inner loop learning rate and a hyperparameter 
        params =  params - alpha*0.5*(Binary.apply(m)+1)*grads
\end{lstlisting}

\end{document}